\documentclass{article}

\usepackage{microtype}
\usepackage{graphicx}
\usepackage{subfigure}
\usepackage{fullpage}
\usepackage[colorlinks,linkcolor=blue,citecolor=blue]{hyperref}
\usepackage{booktabs} 

\usepackage{hyperref}

\newcommand{\AlgA}{\mathcal{A}}
\newcommand{\AlgM}{\mathcal{M}}
\newcommand{\AlgD}{\mathcal{D}}
\newcommand{\PlotWidth}{0.7\linewidth}
\newcommand{\BigPlotWidth}{\linewidth}
\newcommand{\SmallPlotWidth}{0.5\linewidth}
\newcommand{\AppendixPlotWidth}{0.7\linewidth}

\usepackage{amsmath,amsfonts,amssymb}
\usepackage{mathtools}
\usepackage{amsthm} 
\usepackage{latexsym}
\usepackage{relsize}
\usepackage{xspace}
\usepackage{bm}

\usepackage[capitalise]{cleveref}
\usepackage{xcolor}
\usepackage{dsfont}

\newcommand{\defeq}{\stackrel{\text{def}}{=}}

\usepackage{algorithm}
\usepackage{algorithmic}

\newcommand{\D}{\mathcal{D}}

\newcommand{\ignore}[1]{}

\theoremstyle{plain}
\newtheorem{theorem}{Theorem}

\newtheorem{proposition}[theorem]{Proposition}

\newtheorem*{theorem*}{Theorem}
\newtheorem*{lemma*}{Lemma}
\newtheorem*{corollary*}{Corollary}
\newtheorem*{proposition*}{Proposition}
\newtheorem*{claim*}{Claim}
\newtheorem*{fact*}{Fact}
\newtheorem*{observation*}{Observation}

\theoremstyle{definition}

\newtheorem*{definition*}{Definition}
\newtheorem*{remark*}{Remark}
\newtheorem*{example*}{Example}

 \theoremstyle{plain}
\newtheorem*{theoremaux}{\theoremauxref}
\gdef\theoremauxref{1}

\DeclareMathAlphabet{\mathbfsf}{\encodingdefault}{\sfdefault}{bx}{n}

\DeclareMathOperator*{\argmin}{arg\,min}

\newcommand{\norm}[1]{\|#1\|}

\newcommand{\E}{\mathbb{E}}

\newcommand{\tr}{^{\mkern-1.5mu\mathsf{T}}}

\newcommand{\reals}{\mathbb{R}}
\newcommand{\eps}{\varepsilon}

\renewcommand{\leq}{~\le~}

\let\oldtfrac\tfrac
\renewcommand{\tfrac}[2]{\smash{\oldtfrac{#1}{#2}}}

\let\nablaold\nabla
\renewcommand{\nabla}{\nablaold\mkern-2.5mu}

\newcommand{\pa}[1]{\left(#1\right)}
\newcommand{\ang}[1]{\left<#1\right>}

\usepackage{amsmath,amsfonts,bm}

\def\eqref#1{equation~\ref{#1}}

\def\1{\bm{1}}

\def\eps{{\epsilon}}

\DeclareMathAlphabet{\mathsfit}{\encodingdefault}{\sfdefault}{m}{sl}
\SetMathAlphabet{\mathsfit}{bold}{\encodingdefault}{\sfdefault}{bx}{n}

\DeclareMathOperator*{\Minimize}{\mathrm{minimize}}

\newcommand{\citet}[1]{\cite{#1}}
\newcommand{\citep}[1]{\cite{#1}}
\newcommand{\citealt}[1]{\cite{#1}}

\usepackage[italic,defaultmathsizes]{mathastext}

\title{Disentangling Adaptive Gradient Methods from Learning Rates}

\author{
  Naman Agarwal$^1$ \qquad Rohan Anil$^2$ \qquad Elad Hazan$^{1 3}$ \\
  Tomer Koren$^4$ \qquad Cyril Zhang$^{1 3}$ \\
  \\
  $^1$ Google AI Princeton \qquad
  $^2$ Google Brain \\
  $^3$ Department of Computer Science, Princeton University \\
  $^4$ Tel Aviv University \& Google Research, Tel Aviv \\
  \\
  \texttt{\{rohananil, namanagarwal\}@google.com,} \\
  \texttt{\{ehazan, cyril.zhang\}@cs.princeton.edu}, \\
  \texttt{tkoren@tauex.tau.ac.il}
}

\begin{document}
\maketitle

\begin{abstract}
We investigate several confounding factors in the evaluation of optimization algorithms for deep learning. Primarily, we take a deeper look at how adaptive gradient methods interact with the \emph{learning rate schedule}, a notoriously difficult-to-tune hyperparameter which has dramatic effects on the convergence and generalization of neural network training. We introduce a  ``grafting'' experiment which decouples an update's magnitude from its direction, finding that many existing beliefs in the literature may have arisen from insufficient isolation of the implicit schedule of step sizes. Alongside this contribution, we present some empirical and theoretical retrospectives on the generalization of adaptive gradient methods, aimed at bringing more clarity to this space.

\end{abstract}

\section{Introduction}
Adaptive gradient methods are a cornerstone of optimization for deep learning. However, precise characterization of their behavior proves to be elusive; theoretical and empirical accounts of their preconditioning, generalization, and auto-tuning properties have seldom resulted in the unanimous adoption of new practices. In the pursuit of state-of-the-art results, the choice of optimizer is essentially treated as a black-box hyperparameter. To make matters worse, each optimizer implicitly specifies a different implicit sequence of step sizes; this interferes with the tuning of another notoriously opaque hyperparameter, the learning rate schedule.

In this work, we present methods to decouple an optimizer's preconditioning effect from its implicit step size schedule. The chief purpose of this contribution is to provide a more rigorous way to conduct research in optimizer evaluation, but we explore the diagnostic and algorithmic potential of mixing optimizers. In the final section, we provide a broader retrospective on existing beliefs about adaptive gradient methods, and point out instances where experimental conclusions could have arisen from insufficient disentanglement of the optimizer from the learning rate schedule.

\subsection{Our contributions}
To better understand how adaptive gradient methods interact with learning rates, we present the following:

\paragraph{Methodology of learning-rate grafting.} We propose several variants of a simple \emph{grafting} experiment, which combines the step magnitude and direction of two different optimizers. An immediate application is to control for the high-dimensional learning rate schedule parameter when evaluating any optimizer's performance. We also highlight the potential of grafting to enable the discovery of new algorithmic tools. In particular, we find a \emph{linearly increasing} step size schedule correction for AdaGrad.
\paragraph{Empirical insights about preconditioning.} Applying our grafting meta-optimizer to large-scale problems as a diagnostic tool, we find that the \emph{foremost} determining factor of an optimizer's training curve is its implicit step size schedule. Controlling for this enables us to take a deeper look at the individual role of adaptivity. We shed light on why optimizer choice appears to be brittler in natural language processing than in computer vision.
\paragraph{Review of oft-cited evidence.} We provide a broader discussion on the existing literature on the comparison and evaluation of adaptive optimizers. We identify a few common confounding factors, along with easy ways to mitigate them. Centrally, we provide a retrospective on theoretical and empirical evidence from \citet{wilson2017marginal}, including a replication study of their experiments.

Although the core of our work comprises a relatively simple experimental idea, the discussion and implications are quite wide-ranging. For convenience, we collect in Appendix~\ref{appendix:experiment-guide} a short summary all of our empirical findings.

\subsection{Related work} \label{subsec:related}

\paragraph{Adaptive regularization.} Adaptive gradient methods were introduced by the online learning community \citep{duchi2011adaptive,mcmahan2010adaptive}; see \citep{hazan2016introduction} for an introduction. A flurry of extensions, heuristics and modifications followed, most notably RMSprop~\citep{tieleman2012lecture} and Adam~\citep{kingma2014adam}.

\paragraph{Adaptive optimization in deep learning.} Adaptive methods have turned out to be extremely robust in training deep neural networks, receiving tens of thousands of citations for this reason. In particular, Adam has been the \emph{de facto} standard in fields such as NLP \citep{devlin18,yang2019xlnet,liu2019roberta}, deep generative modeling \citep{karras2017progressive,brock2018large,kingma2018glow}, and deep RL \citep{haarnoja2018soft}. Adaptive methods have seen adoption in extremely large-scale settings, necessitating modifications to reduce resource consumption \citep{shazeer2018adafactor, anil2019memory, chen2019extreme}.

\paragraph{The empirical debate on adaptive methods.}
An important discussion was sparked by \citet{wilson2017marginal}, who presented empirical and theoretical situations where adaptive methods generalize poorly. Building on this premise, \citet{keskar2017improving} suggest switching from Adam and SGD during training. \citet{smith2019super} develop a doctrine of ``superconvergence'' which eschews adaptive methods. \citet{reddi2018convergence} construct, then mitigate, a pathological setting where Adam fails to converge. However, in the vast majority of cases, out-of-the-box adaptive methods are perfectly suitable for practitioners.

\paragraph{Learning rate schedules.} Choosing learning rate schedules has become a vexing empirical problem. Beyond classical optimization, \citet{ge2019step} provide a fine-grained theoretical account for quadratic losses; \cite{li2019exponential,arora2018theoretical} study the interaction of learning rates with batch normalization \cite{ioffe2015batch}. The interaction of learning rates with batch size is explored in \citet{krizhevsky2014one,goyal2017accurate,bottou2018optimization}. Learning rate warmups and restarts are popular state-of-the-art heuristics \citep{gotmare2018closer,loshchilov2016sgdr}. \citet{shampoopp} use the grafting method as one of many heuristics to stabilize an exotic full-matrix optimizer; we comment further in Section~\ref{subsec:implications}.

\paragraph{Discarding the backprop magnitude.} Several recent lines of work propose optimization procedures which discard magnitude information from the gradients (without replacing it with another optimizer's magnitude). \citep{blier2018learning} train models with randomized learning rates, motivated by ensemble learning intuitions. See the discussion on exotic optimizers in Section~\ref{subsec:implications} for more examples. Our grafting meta-algorithm might help to shed more light on why these optimizers converge successfully.
\section{Preliminaries}

\subsection{Stochastic first-order optimization}
We will work in the usual abstraction of the \emph{stochastic optimization} problem:
\begin{align*}
    \Minimize_{w \in \mathcal{W}}
    \;\;
    F(w) \defeq \E_{z \sim \D}[f(w;z)],
\end{align*}
where the expectation is over a random variable $z$ whose distribution $\D$ is initially unknown. In supervised learning, $z$ represents an example-label pair $(x,y)$, drawn from an unknown population.
A stochastic optimization algorithm is given samples $z_1,\ldots,z_T \sim \D$ from the underlying distribution, and produces a point $\bar{w} \in \reals^d$ whose population loss $F(\bar{w})$ is as close as possible to that of the minimizer $w^\star = \argmin_w F(w)$. We will consider domains $\mathcal{W} \subseteq \reals^d$.

We will focus exclusively on methods which maintain a sequence of iterates $(w_1,\ldots,w_T)$, iteratively querying a \emph{stochastic gradient} $g_t$, for which $\E[g_t] = \nabla F(w_t)$. In modern deep learning, this is typically $\nabla f(w_t;z)$ averaged over a mini-batch. Notationally, a \emph{gradient optimizer} is a map from the current iterate and gradient $(w_t, g_t)$ to the next iterate $w_{t+1}$.\footnote{Note that our notation is ``imperative'' rather than ``functional'': an update can depend on the algorithm's internal state, hyperparameters, or randomness.} Then, vanilla SGD with learning rate schedule $\eta_t$ is the algorithm $\AlgA$ specified by 
\[\AlgA(w_t, g_t) := w_t - \eta_t g_t.\]

\subsection{Adaptive gradient methods} \label{subsec:adaptive-gradients}

It will be convenient to introduce some unifying notation for the standard family of second-moment-based adaptive optimizers. Algorithm~\ref{alg:generic-grad} captures the usual formulations of Adam, AdaGrad, and RMSprop, up to scaling conventions which can be absorbed into $\{\eta_t\}$; see Section~\ref{subsec:software} for a discussion on these, and \citet{choi2019empirical} for a comprehensive unifying treatment. Additionally, by omitting the denominator $\sqrt{m_2 + \eps}$, we recover SGD with momentum. All operations on vectors are understood to be entrywise; $(\cdot)^{\odot 2}$ indicates entrywise squaring.

\begin{algorithm} 
\caption{ Generic second-moment adaptive optimizer }
\label{alg:generic-grad}
\begin{algorithmic}[1]
\STATE \textbf{Input: } Initializer $x_1$, learning rate schedule $\{\eta_t\}$, hyperparameters $(\beta_1, \beta_2, \eps)$.
\STATE Initialize $m_1, m_2 \in \reals^d$.
\FOR{$t = 1, \ldots, T$}
  \STATE Receive stochastic gradient $g_t$ at $x_t$.
  \STATE Update accumulators:
  \[ m_1 \leftarrow \beta_1 m_1 + g_t , \quad m_2 \leftarrow \beta_2 m_2 + g_t^{\odot 2}. \]
  \STATE Update:
  \[ x_{t+1} \leftarrow x_t + \frac{\eta_t}{ \sqrt{m_2 + \eps} } \cdot m_1. \]
\ENDFOR
\end{algorithmic}
\end{algorithm}

The exponential window parameter $\beta_1$ denotes heavy-ball momentum \citep{polyak1964some}. We set this to be $0.9$ throughout our experiments, which remains an uncontroversial rule of thumb. Importantly, we introduce momentum in AdaGrad, even though this is not possible by default in most deep learning packages (see Section~\ref{subsec:software}). We \emph{do not} attempt to disentangle the role of momentum in deep learning, nor the Nesterov-style alternative update \citep{nesterov1983method,dozat2016incorporating}; we view this as an important but much subtler orthogonal question.

Crucially, for AdaGrad we have $\beta_2 = 1$; the accumulator does not decay. Adam, RMSprop, and countless variants opt for a slow attenuation like $\beta_2 = 0.999$, as we use for all Adam experiments. The role of this is controversial; some works \citep{kingma2014adam,agarwal2017second,staib2019escaping} view this a feature (forgetting stale information about the loss surface's curvature), while \citet{reddi2018convergence} present pathological failure cases that arise from this same forgetfulness. One minor conclusion from our main empirical study (Section~\ref{sec:experiments}) is that AdaGrad's accumulator rule is not necessarily harmful in deep learning.

The constant $\eps$ is used to promote numerical stability in low-precision division, but can also be interpreted as a way to \emph{discard} adaptivity and ``interpolate'' an algorithm with SGD. We discuss the confounding perils of this in Section~\ref{subsec:software}, and use $\eps = 0$ in our experiments, as outlined there.

\subsection{Step size schedules, implicit and explicit}
It is well-known that adaptive methods imply effective step sizes, especially in light of the discussion on the second-moment accumulator. This is motivated by the online formulation, in which a $1/\sqrt{t}$ decay of the update magnitude gives optimal worst-case regret bounds \citep{zinkevich2003online,duchi2011adaptive}. Going even earlier, step-size sequences are needed for classical convergence guarantees in first-order non-smooth and/or stochastic optimization (see, e.g., \citealt{boyd2004convex,ghadimi2013stochastic}).
Table~\ref{table:theoretical-rates} gives a survey of some theoretically and/or empirically-motivated step size schedules.

\begin{table}
\centering
\begin{tabular}{|c|c|}
\hline
\textbf{\quad Schedule \quad} & \textbf{\quad Setting \quad} \\ \hline
constant & \begin{tabular}{c}
     Smooth, convex GD \\ \cite{nesterov2013introductory} 
\end{tabular}
 \\ \hline
$t^{-1}$ & \begin{tabular}{c}
     Strongly convex SGD \\ \cite{hazan2011beyond} 
\end{tabular} \\ \hline
$t^{-1/2}$ &  \begin{tabular}{c}
     Smooth non-convex SGD \\ \cite{ghadimi2013stochastic}; \\
     Non-smooth convex GD/SGD \\
     \cite{nesterov2013introductory}
\end{tabular}\\ \hline
$c^{-t}$ (decaying) & \begin{tabular}{c}
     SGD, least squares \\\cite{ge2019step} 
\end{tabular}  \\ \hline
$c^t$ (growing) & \begin{tabular}{c}
     Batch norm + $\ell_2$ weight decay \\ \cite{li2019exponential} 
\end{tabular}  \\ \hline
$\cos(c_0 + c_1 t)$ & \begin{tabular}{c}
     Image models \\  \cite{loshchilov2016sgdr} 
\end{tabular}\\
\hline
\end{tabular}
\label{table:theoretical-rates}
\caption{Some known learning rate schedules from classical and recent optimization literature.}
\end{table}
However, a general theory of how to set the learning rate schedule is notoriously elusive. This $T$-dimensional hyperparameter is generally selected by human search, augmented by architecture-specific heuristics.\footnote{It is difficult to find an authoritative citation for this, but refer to the step size heuristics in \citet{goyal2017accurate,popel2018training} to see how delicate they are in state-of-the-art pipelines.} For example, stepwise exponentially-decaying schedules are very popular on vision tasks, but seldom seen on language tasks. To add further complexity, learning rate warmups have been a recent popular choice across domains.  

\section{Decoupling Step Size from Direction} \label{sec:experiments}
Concretely, we point out the following tension in the practice of optimization in deep learning: that the role of $\eta_t$ overlaps with the \emph{implicit step size schedule} induced by the optimizer. In this section, we propose decoupling these factors via \emph{learning rate grafting}.

\subsection{AdaGraft: a meta-optimizer}

We begin by presenting our experimental methodology, which revolves around Algorithm~\ref{alg:adagraft}. At each iteration, it computes a single gradient, passes it to two optimizers, and makes a \emph{grafted} step, combining the \emph{magnitude} of $\AlgM$'s step and \emph{direction} of $\AlgD$'s step.

\begin{algorithm} 
\caption{ AdaGraft meta-optimizer }
\label{alg:adagraft}
\begin{algorithmic}[1]
\STATE \textbf{Input: } Optimizers $\AlgM, \AlgD$; initializer $w_1$; $\eps > 0$.
\STATE Initialize $\AlgM, \AlgD$ at $w_1$.
\FOR{$t = 1, \ldots, T$}
  \STATE Receive stochastic gradient $g_t$ at $w_t$.
  \STATE Query steps from $\AlgM$ and $\AlgD$:
  \[w_\AlgM := \AlgM(w_t, g_t), \quad w_\AlgD := \AlgD(w_t, g_t).\]
  \STATE Update with grafted step:
  \[ w_{t+1} \leftarrow w_t + \frac{\norm{w_\AlgM - w_t}}{\norm{w_\AlgD - w_t} + \eps} \cdot \pa{ w_\AlgD - w_t }. \]
\ENDFOR
\end{algorithmic}
\end{algorithm}

\paragraph{Layer-wise vs. global grafting.} Two natural variants of Algorithm~\ref{alg:adagraft} come to mind, especially if one is concerned about efficient implementation (see Appendix~\ref{subsec:implementation}). In the \emph{layer-wise} version, we view $w_t$ as a single parameter group (usually a tensor-shaped variable specified by the architecture), and apply AdaGraft and its child optimizers to each group. In the \emph{global} version, $w_t$ contains all of the model's weights. We discuss and evaluate both variants, but our main experimental results use \emph{layer-wise} grafting.

The first experimental question addressed by AdaGraft is the following: \emph{To what degree does an optimizer's implicit step size schedule determine its training curve?} To this end, given a set of base optimizers, we can perform training runs for all pairs $(\AlgM, \AlgD)$, where grafting $(\AlgA, \AlgA)$ is understood as simply running $\AlgA$. For the main experiments, we use SGD, Adam, and AdaGrad, all with momentum $\beta_1 = 0.9$.

All experiments were carried out on 32 cores of a TPU-v3 Pod \citep{jouppi2017datacenter}, using the Lingvo \citep{shen2019lingvo} sequence-to-sequence framework built on top of TensorFlow \citep{abadi2016tensorflow}.

\subsection{ImageNet classification experiments} \label{subsec:imagenet}
We ran all pairs of grafted optimizers on a 50-layer residual network \citep{he2016identity} with 26M parameters, trained on ImageNet classification \citep{imagenet}. We used a batch size of 4096, enabled by the large-scale training infrastructure, and a learning rate schedule consisting of a linear warmup and stepwise exponential decay. All details can be found in Appendix~\ref{subsec:imagenet-details}.

\begin{figure}
    \centering
    \includegraphics[width=\PlotWidth]{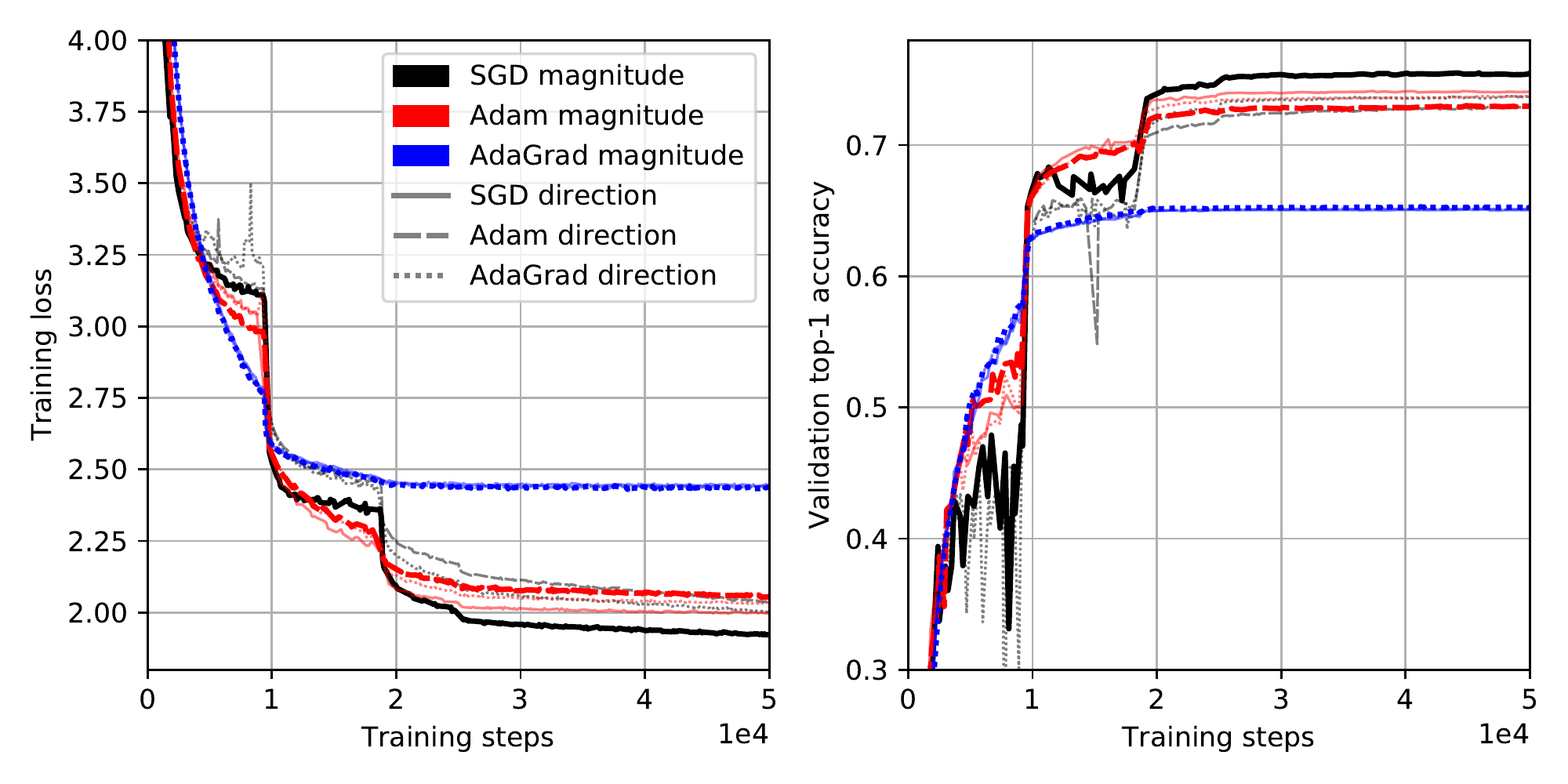}
    \caption{Training curves for grafted optimizers on ResNet-50 for ImageNet classification. To first order, the shape of the training curve is determined by the step size schedule; most dramatically, \textcolor{blue}{AdaGrad}'s poor convergence appears to be entirely due to its implicit step size schedule.}
    \label{fig:imagenet-graft}
\end{figure}

Table~\ref{table:imagenet-results} shows top-1 and top-5 accuracies at convergence. The final accuracies at convergence, as well as training loss curves, are very stable ($<0.1\%$ deviation) across runs, due to the large batch size. Figure~\ref{fig:imagenet-graft} shows at a glance our main empirical observation: that the shapes of the training curves are clustered by the choice of $\AlgM$, the optimizer which supplies the step magnitude.

We stress that no additional hyperparameter tuning was done in these experiments; not even the global scalar learning rate needed adjustment. Thus, starting with $N$ tuned optimizer setups, grafting produces a table of $N^2$ setups with no additional effort. Each row of this table controls for the implicit step size schedule.

\newcommand{\sep}{\textcolor{gray}{$\;\cdot\;$} }

\begin{table}[H]
\centering
\begin{tabular}{|c|c|c|c|}
\hline
$\bm{\AlgM \; \diagdown \; \AlgD}$ & \textbf{SGD} & \textbf{Adam} & \textbf{AdaGrad} \\ \hline
\textbf{SGD} & 75.4 \sep 92.6 & 72.8 \sep 91.2 & 73.7 \sep 91.4 \\ \hline
\textbf{Adam} & 74.1 \sep 91.9 & 73.0 \sep 91.3 & 73.7 \sep 91.6 \\ \hline
\textbf{AdaGrad} & 65.0 \sep 85.9 & 65.1 \sep 86.0 & 65.3 \sep 86.3 \\ \hline
\end{tabular}
\caption{Top-1 and top-5 accuracies at training step $t = 50$K for ImageNet experiments. Averaged over 3 trials; no accuracy varied by more than $0.1\%$.}
\label{table:imagenet-results}
\end{table}

\subsection{WMT14 English-French translation experiments} \label{subsec:wmt}
For a realistic large-scale NLP setting, we trained all grafted optimizers on a 6-layer Transformer network \citep{vaswani2017attention} with 375M parameters, on the WMT14 English-French translation task, which has 36.3M sentence pairs. Again, we use a large batch size (384 sequences), enabling very robust training setups. 
More details can be found in Appendix~\ref{subsec:wmt-details}.

\begin{figure}
    \centering
    \includegraphics[width=\PlotWidth]{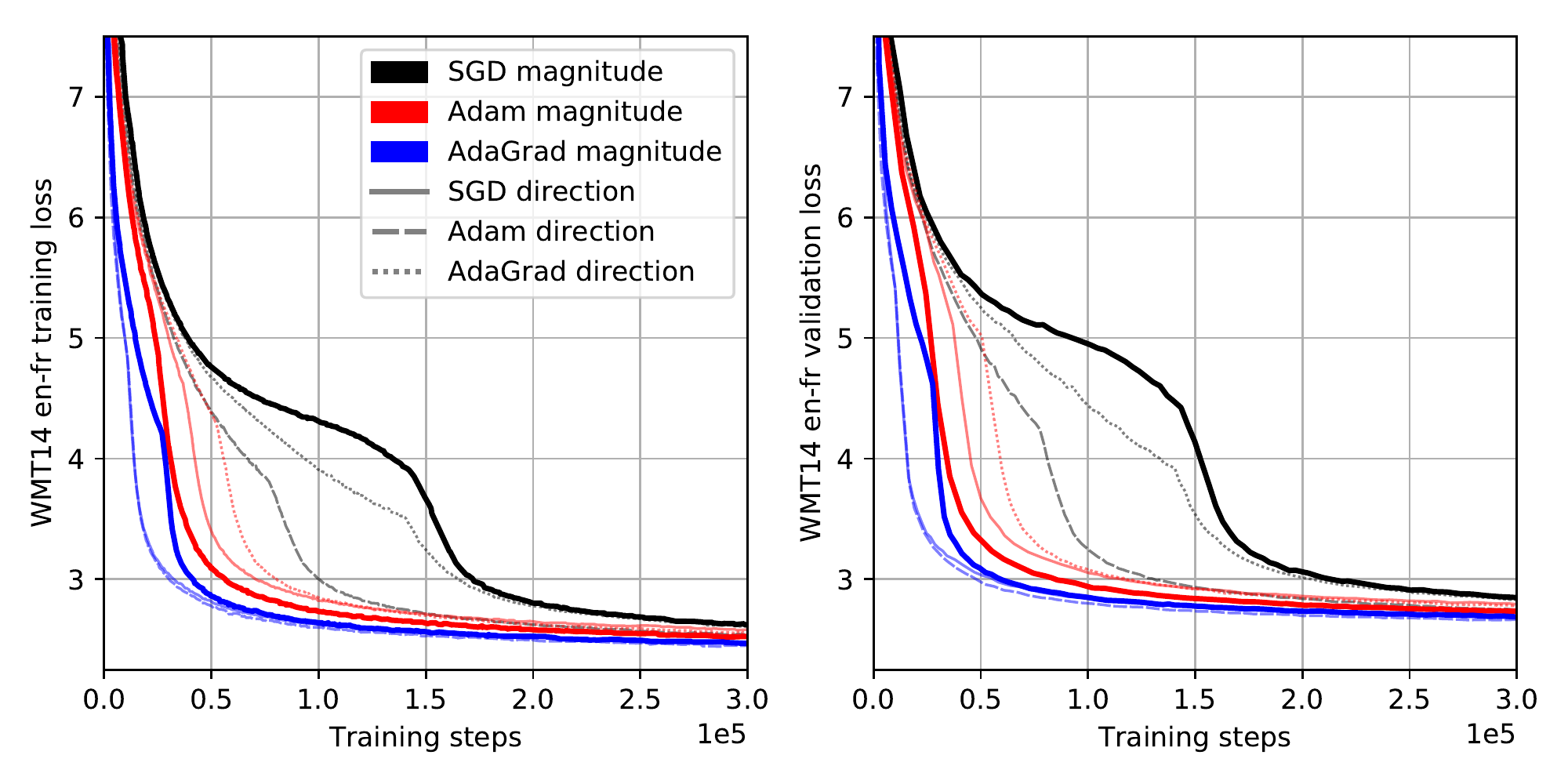}
    \caption{Training curves for grafted optimizers on Transformer for WMT14 English-French translation. Convergence and solution quality are very sensitive to implicit step size schedule.}
    \label{fig:wmt-graft}
\end{figure}

\begin{table}[H]
\centering
\begin{tabular}{|c|c|c|c|}
\hline
$\bm{\AlgM \; \diagdown \; \AlgD}$ & \textbf{SGD} & \textbf{Adam} & \textbf{AdaGrad} \\ \hline
\textbf{SGD} &     39.8 $\pm$ 0.1 & 40.0 $\pm$ 0.3 & 39.4 $\pm$ 0.2 \\ \hline
\textbf{Adam} &    40.6 $\pm$ 0.2 & 41.2 $\pm$ 0.2 & 40.1 $\pm$ 0.2 \\ \hline
\textbf{AdaGrad} & 41.4 $\pm$ 0.2 & 41.8 $\pm$ 0.1 & 41.6 $\pm$ 0.1 \\ \hline
\end{tabular}
\label{table:wmt-results}
\caption{Test set BLEU scores for Transformer grafting experiments at $t = 300$K training steps. Averaged over 3 trials; standard deviations shown.}
\end{table}

Interestingly, beyond demonstrating the same clustering of performance metrics by the choice of $\AlgM$, these experiments show that it is possible for a grafted optimizer to outperform both base methods $\AlgM$ and $\AlgD$; see Figure~\ref{fig:wmt-graft} for loss curves, and Table~\ref{table:wmt-results} for the downstream BLEU metric, with which our results are consistent. Again, we are not making claims of categorical superiority under careful tuning, and only the power of bootstrapping; we stress that we did not even tune the global learning rate scalar.

\subsection{Bootstrapping a learning rate schedule} \label{subsec:global-grafting}
So far, Algorithm~\ref{alg:adagraft} has taken the role of an exploratory tool for more principled optimizer comparison. In this section, we illustrate how it can be used in optimizer discovery. Focusing on the underrepresentation of AdaGrad in computer vision, we see that grafting can be repurposed to find a suitable learning rate schedule from scratch, which might not have been found from blind search or first principles.

It has been noted many times \citep{zeiler2012adadelta,wilson2017marginal,bottou2018optimization} that AdaGrad's step size schedule is ``too aggressive'' for deep learning. Using the \emph{global} variant of Algorithm~\ref{alg:adagraft} with $(\AlgM,\AlgD)$ = (SGD, AdaGrad) and keeping track of the norms of the steps produced by $\AlgM$ and $\AlgD$ (Figure~\ref{fig:imagenet-stepnorms}), we arrive at a \emph{linear} correction $\{ \eta_t = c_0 + c_1 t \}$, with $(c_0, c_1) \approx (0.2,10^{-4})$, which can be overlaid on the usual warmup and stepwise-exponential annealing.

\begin{figure}
    \centering
    \includegraphics[width=\PlotWidth]{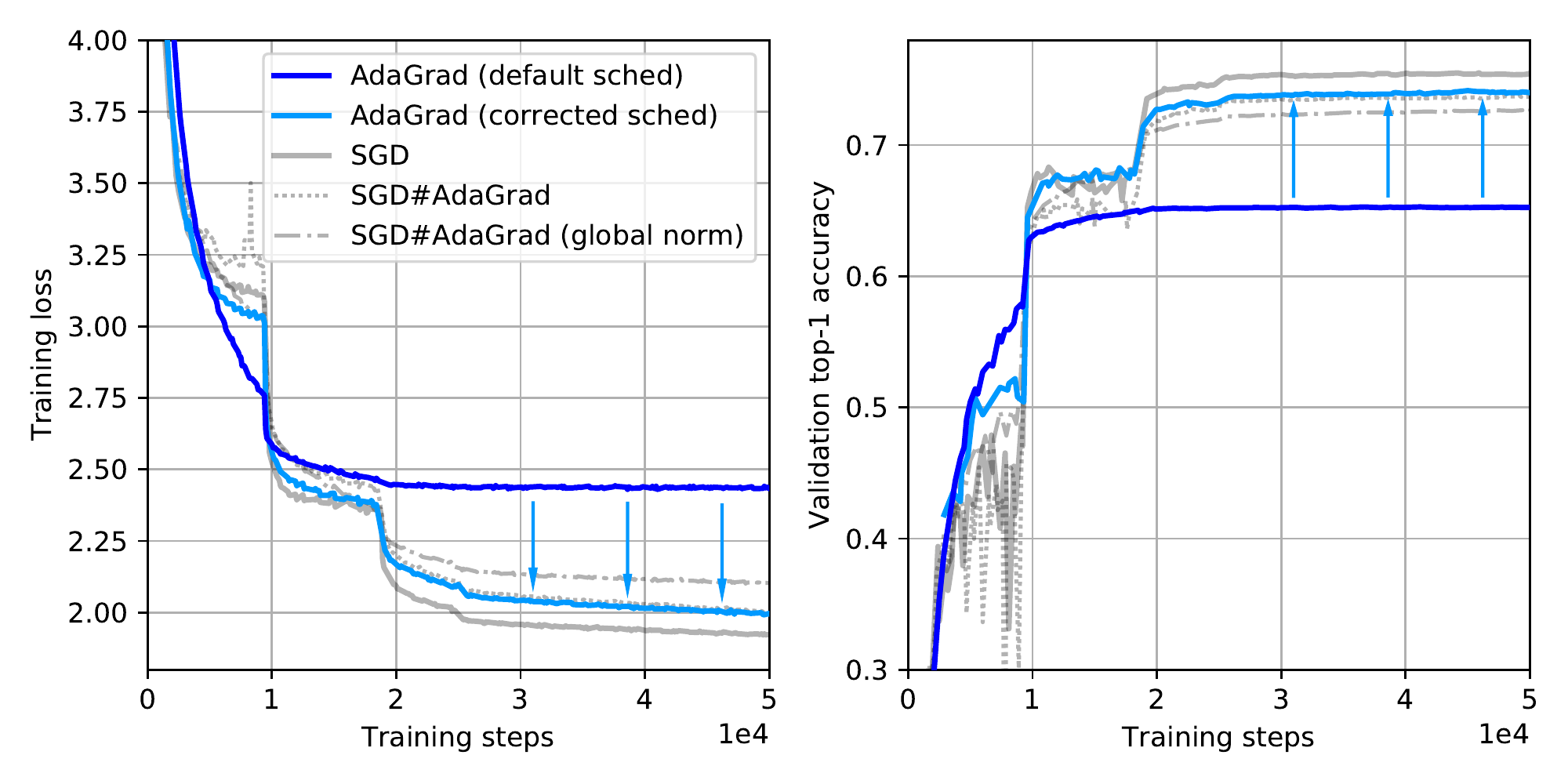}
    \caption{Global grafting experiments to fix AdaGrad. Bootstrapping with a well-tuned $\AlgM$=SGD, we find an \emph{offline} correction for \textcolor{blue}{AdaGrad}'s implicit step size schedule; \textcolor[rgb]{0,0.6,1}{corrected AdaGrad} competes with SGD or Adam. Grafting is denoted by $\AlgM\#\AlgD$.}
    \label{fig:imagenet-transfer}
\end{figure}

\begin{figure}
    \centering
    \includegraphics[width=\PlotWidth]{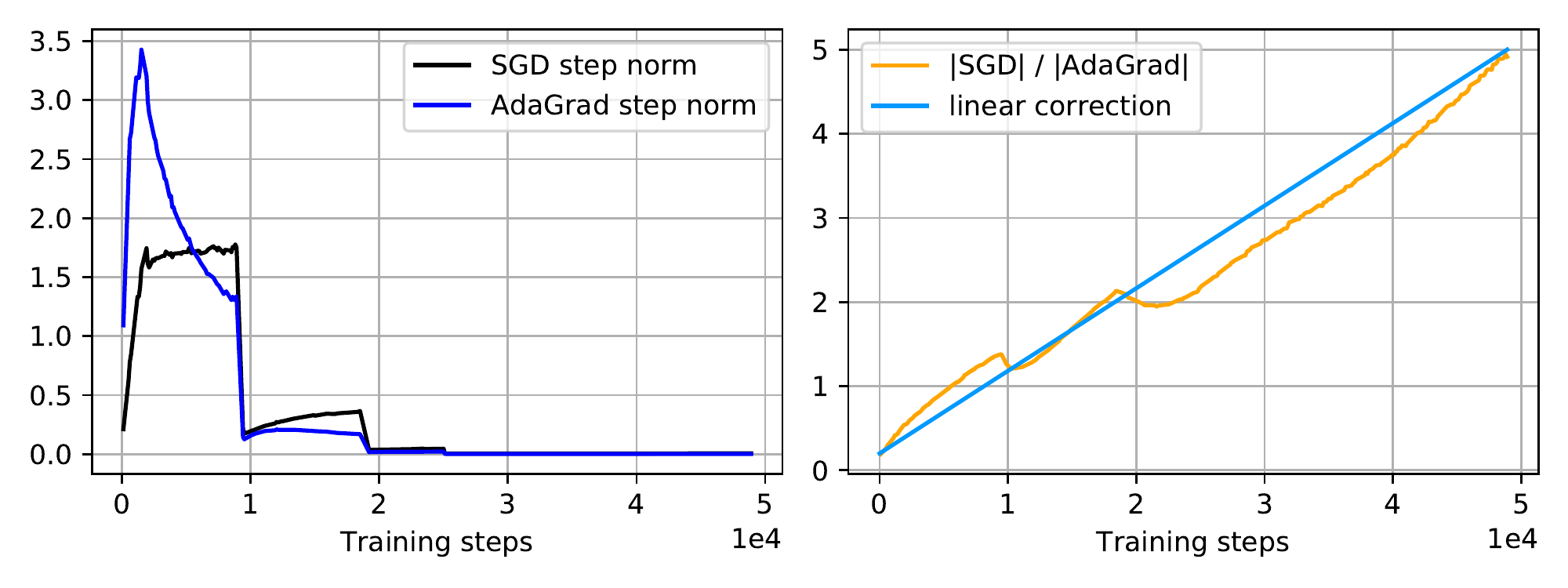}
    \caption{Empirical discovery of linear learning rate correction from Figure~\ref{fig:imagenet-transfer}. Norms of proposed steps are measured in a global (SGD, AdaGrad) grafting experiment. The ratio is found empirically to scale linearly with $t$; this adjustment can be applied directly to AdaGrad, without any grafting.}
    \label{fig:imagenet-stepnorms}
\end{figure}

We have not encountered a linearly increasing learning rate schedule anywhere in the literature (see Table~\ref{table:theoretical-rates}), and its interaction with AdaGrad's $1/\sqrt{t}$ decay is unintuitive. We try this again in the CIFAR-10 experiment in Section~\ref{subsec:replication}, but leave a closer look for future work.

\begin{figure}
    \centering
    \includegraphics[width=\PlotWidth]{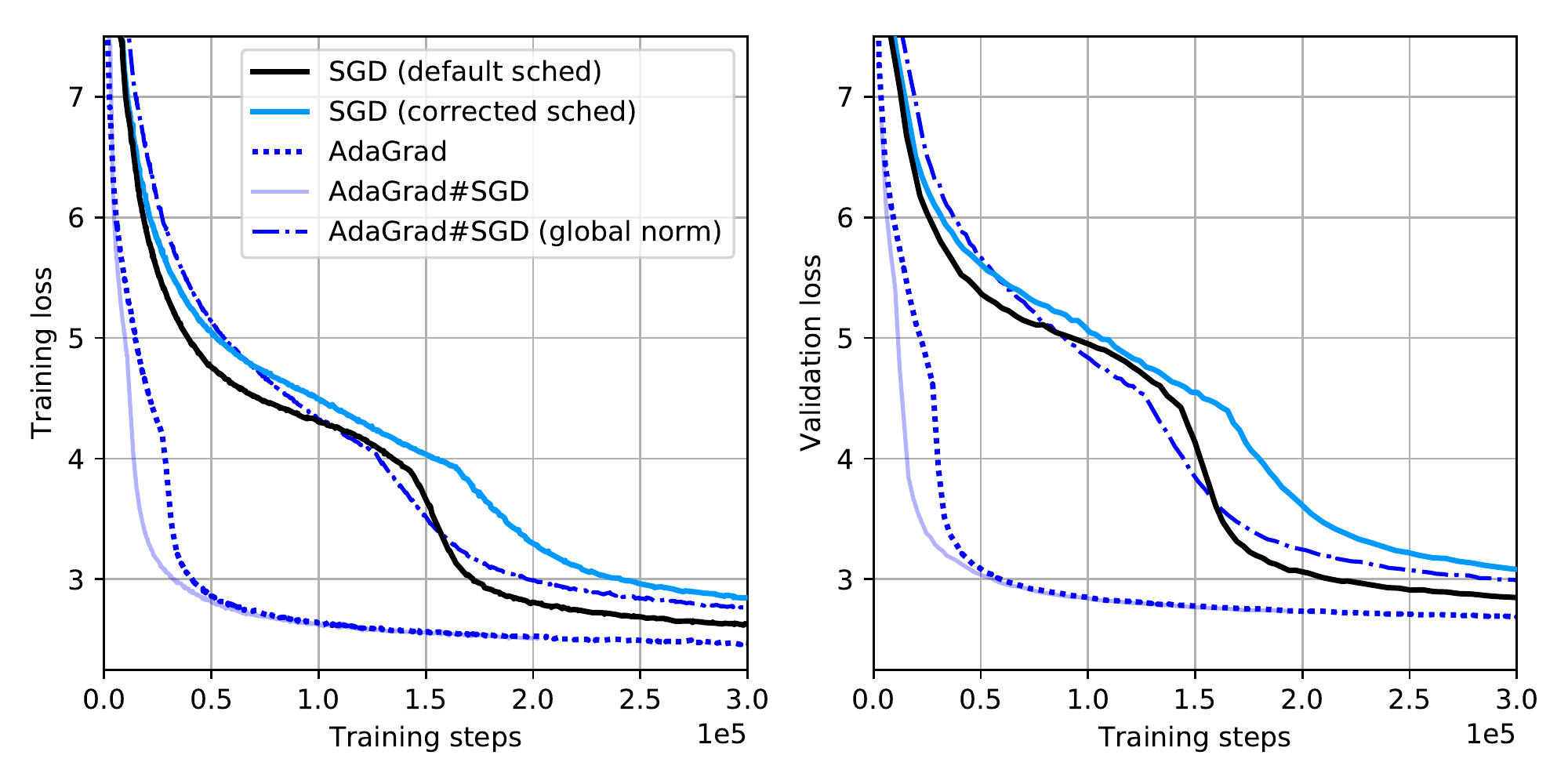}
    \caption{Training curves for global grafting and learning rate schedule correction, as in Figure~\ref{fig:imagenet-transfer}, for the Transformer/WMT setup, layer-wise grafting with $\AlgM$=AdaGrad, $\AlgD$=SGD. Unlike the analogous experiment for ImageNet, we could not get corrected SGD to perform as well as AdaGrad.}
    \label{fig:wmt-transfer}
\end{figure}

\paragraph{Insufficiency of global learning rates in translation.} Repeating this experiment in the machine translation setup, we discover a $t^{-0.27}$ learning rate correction from SGD to AdaGrad using \emph{per-layer} grafting; however, we could not get SGD with this offline correction to train like AdaGrad; see Figure~\ref{fig:wmt-transfer}. Visualizing the per-layer norm corrections prescribed by Algorithm~\ref{alg:adagraft} sheds some light on why this might be the case; see Section~\ref{subsec:nlp} and Appendix~\ref{subsec:wmt-transfer} for further discussion and some additional visualizations.

\section{Discussion and Further Experiments} \label{sec:discussion}
\subsection{Better evaluation for the zoo of optimizers} \label{subsec:implications}

\begin{figure}
    \centering
    \includegraphics[width=\PlotWidth]{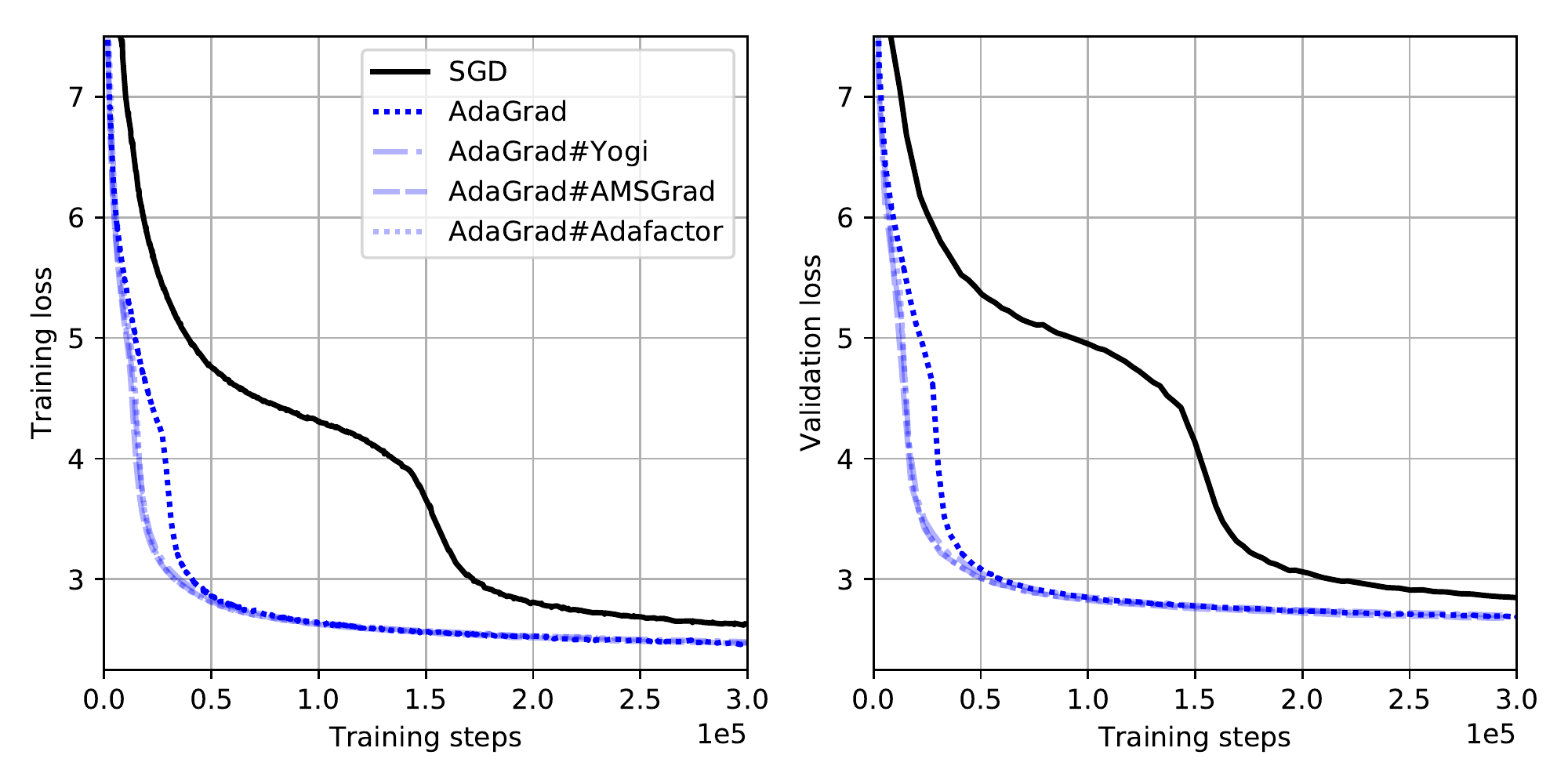}
    \caption{Grafting experiments with an assortment of optimizers, fixing $\AlgM$=AdaGrad. The training curves are nearly indistinguishable for the closely related adaptive methods, which we did not tune at all. }
    \label{fig:wmt-exotic}
\end{figure}

We highlight the potential of grafting to control for the implicit step size schedules of optimizers whose dynamics are not well-understood. In Figure~\ref{fig:wmt-exotic}, we show that by grafting the step size schedule of AdaGrad onto several recently proposed adaptive optimizers \cite{zaheer2018adaptive,reddi2018convergence,shazeer2018adafactor}, showing that their step directions are not significantly different.

Importantly, note that this does \emph{not} invalidate any empirical findings from those works; it only suggests that these optimizers do not propose better directions, in the sense of coordinate-wise step size scalings. Running meaningful experiments with these optimizers playing the role of $\AlgM$ requires exact replications of their setups, which we do not attempt in this work.

We conclude this section with more connections to the recent large-scale optimization literature.

\paragraph{More exotic update directions.} We highlight the possibility of simplifying the problem of searching for a good learning rate schedule when faced with an exotic optimizer. These might arise from quantization when one is concerned with communication efficiency \citep{bernstein2018signsgd,wen2017terngrad}, optimizer search \cite{bello2017neural}, or various theoretically-motivated optimizers which deviate from usual second-moment statistics \cite{chen2018closing}.

\paragraph{Normalizing steps by layer weights.} A string of recent empirical successes in accelerating large-scale neural network training tie the size of the update step to that of the \emph{layer} $\norm{w_t}$, rather than the statistics of the gradients. These include LARS \citep{you2017large}, LAMB \citep{you2019reducing}, and Novograd \citep{ginsburg2019stochastic}. Our experiments, particularly when grafting $\AlgD =$ SGD, synergize with these works: that layer-wise scaling is powerful enough to train many deep learning architectures.

\paragraph{Beyond diagonal-matrix preconditioning.} An important family of optimizers comes from recent efforts to bring full-matrix preconditioning from theoretical origins to state-of-the-art deep learning \cite{martens2015optimizing,martens2018kronecker,gupta2018shampoo,pmlr-v97-agarwal19b}.
As mentioned in \ref{subsec:related}, learning-rate grafting was one of the many heuristics reported by \citet{shampoopp} in order to make Shampoo converge in an industry-scale setting.

\subsection{Understanding the role of adaptivity in NLP} \label{subsec:nlp}
In state-of-the-art methods in natural language processing powered by Transformers or recurrent neural networks, Adam is a versatile go-to choice. We propose using the per-layer grafting experiments, which were successful in both domains, to help explain why global learning rate correction is insufficient for Transformers.

\begin{figure*}
    \centering
    \includegraphics[width=\BigPlotWidth]{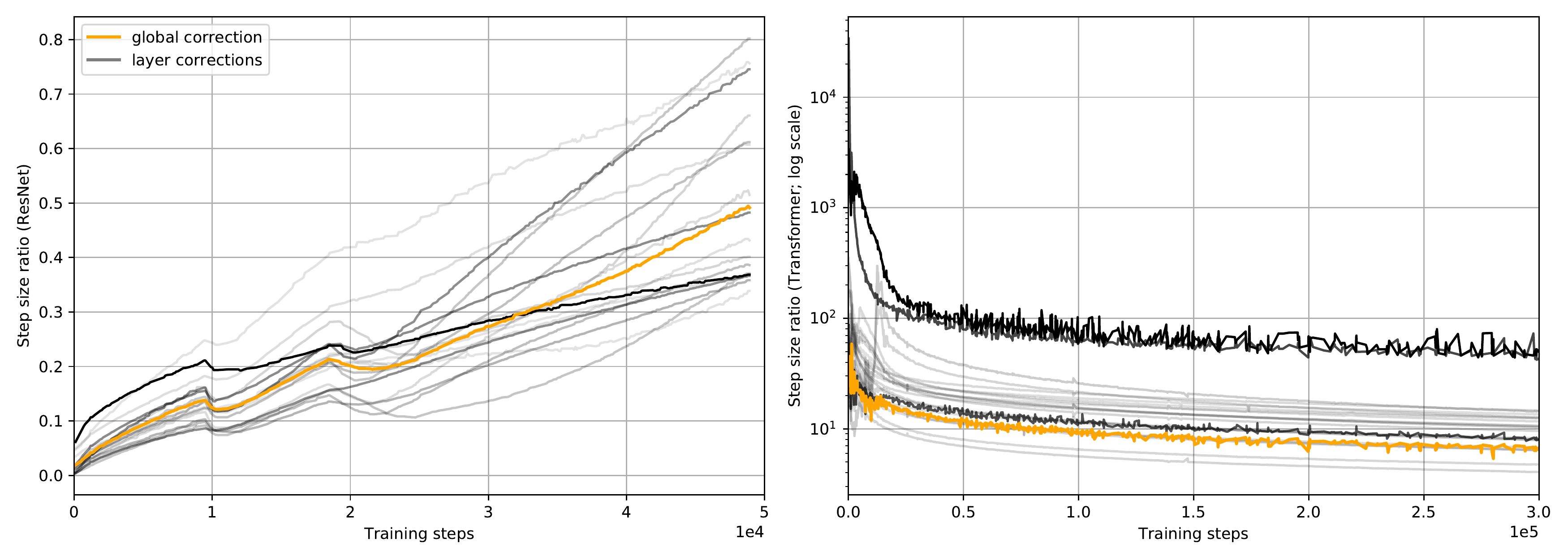}
    \caption{Layer-wise and global step size corrections for grafting experiments from Sections~\ref{subsec:imagenet} and \ref{subsec:wmt}. Curves for \textcolor{gray}{individual layers} are shaded proportionally to their average squared step norm from $\AlgM$. \emph{Left:} ResNet setup, running layer-wise grafting with $\AlgM$ = SGD, $\AlgD$ = AdaGrad.  The \textcolor{orange}{global ratio} is well-correlated with the the individual layers' curves. Darkest curve is the \textbf{softmax} layer. \emph{Right:} Transformer setup, running layer-wise grafting with $\AlgM$ = SGD, $\AlgD$ = AdaGrad. Unlike for ImageNet, the per-layer corrections are many orders of magnitude apart; note that the $y$-axis is \emph{logarithmic}. Darkest curves are the \textbf{softmax} and \textbf{embedding} layers.}
    \label{fig:all-corr}
\end{figure*}

Figure~\ref{fig:all-corr} visualizes the per-layer learning rates used to get successful convergence and matching training curves in the layer-wise grafting experiments. The effect of adaptive methods on Transformer training varies in magnitude and by layer; thus, the trajectory correction is higher-dimensional than on a ResNet, where can be summarized by a time-varying scalar. This provides a finer-grained view of the difficulty of optimization in deep NLP as compared to computer vision \cite{pascanu2013difficulty,pmlr-v97-agarwal19b}.

\section{Revisiting Beliefs about Adaptivity} \label{section:revisiting}
The remainder of this work can be thought of as a ``position paper'': a medley of broader observations on the confounding interactions between adaptive methods and learning rate schedules, arising from the grafting (plus some additional) experiments, as well as recent literature.

\subsection{Theoretical accounts}

\paragraph{Pathological examples.} A frequently-cited justification for the belief that SGD enjoys privileged generalization performance is a pathological case in which adaptive methods overfit dramatically:
\begin{proposition}[\citealt{wilson2017marginal}; informal]
There exists an instance of least-squares regression ($f_i(w; x_i,y_i) = (\ang{w, x_i} - y)^2$) on which SGD generalizes better than AdaGrad.
\end{proposition}
Intuitively, SGD converges to a solution with small $\ell_2$ norm. As is illustrated in the example above this allows SGD to not find a solution in a spurious subspace embedded inside the problem overall making it robust to outliers.

To present a counterpoint to this example, we construct a case in which SGD \emph{underfits}:
\begin{proposition}[Counter-counterexample; informal]
There exists an instance of hinge-loss linear classification ($f_i(w; x_i,y_i) = (1 - y_i \ang{w, x_i})_+$) on which AdaGrad generalizes better than SGD.
\end{proposition}
Intuitively, in this scenario the test error depends on seeing every coordinate and some coordinates are chosen to be \emph{rare}. While AdaGrad adapts quickly, SGD is slower to adapt, leading to comparatively very slow convergence in test error.

Thus, although artificial examples may be illuminating, especially in order to identify edge cases and qualitative understandings for when certain optimizers might fail, one must be careful not to apply these intuitions too generally. We review the first example and present the second in more detail in Appendix~\ref{subsec:pathological-theory}.

\paragraph{Literature on the trajectory analysis of SGD.} Finally, we point to some recent literature on the generalization of SGD, which employ varied theoretical lenses to analyze its optimization trajectory. These include analyses based on stability and early stopping \citep{hardt2015train}, implicit bias in basic settings \citep{ gunasekar2018characterizing, gunasekar2018implicit}, and the specific structure of neural networks \citep{allen2019can, arora2019fine}. Recent work \citep{belkin2019reconciling, mei2019generalization} studies the generalization of gradient descent in the overparameterized \emph{interpolation regime}. These works show that SGD generalizes in various abstractions of neural network training, but do not make negative claims about adaptive methods. Since optimization is so robust to distortion of the gradient signal via preconditioning, our experiments suggest that a useful theoretical account of optimization in deep learning should not be too sensitive to the fact that SGD is the object of interest. It remains an intriguing question whether the trajectories of grafted optimizers match (in some well-chosen metric); we leave a careful experimental probe for future work.

\subsection{Superficial discrepancies between optimizers} \label{subsec:software}
Even though we have taken a unifying view of the usual family of adaptive optimizers, the choice of optimizer is often seen as a black-box discrete hyperparameter. We show here that modulo the learning rate schedule, the popular second-moment adaptive optimizers are equivalent.

\begin{figure}
    \centering
    \includegraphics[width=\SmallPlotWidth]{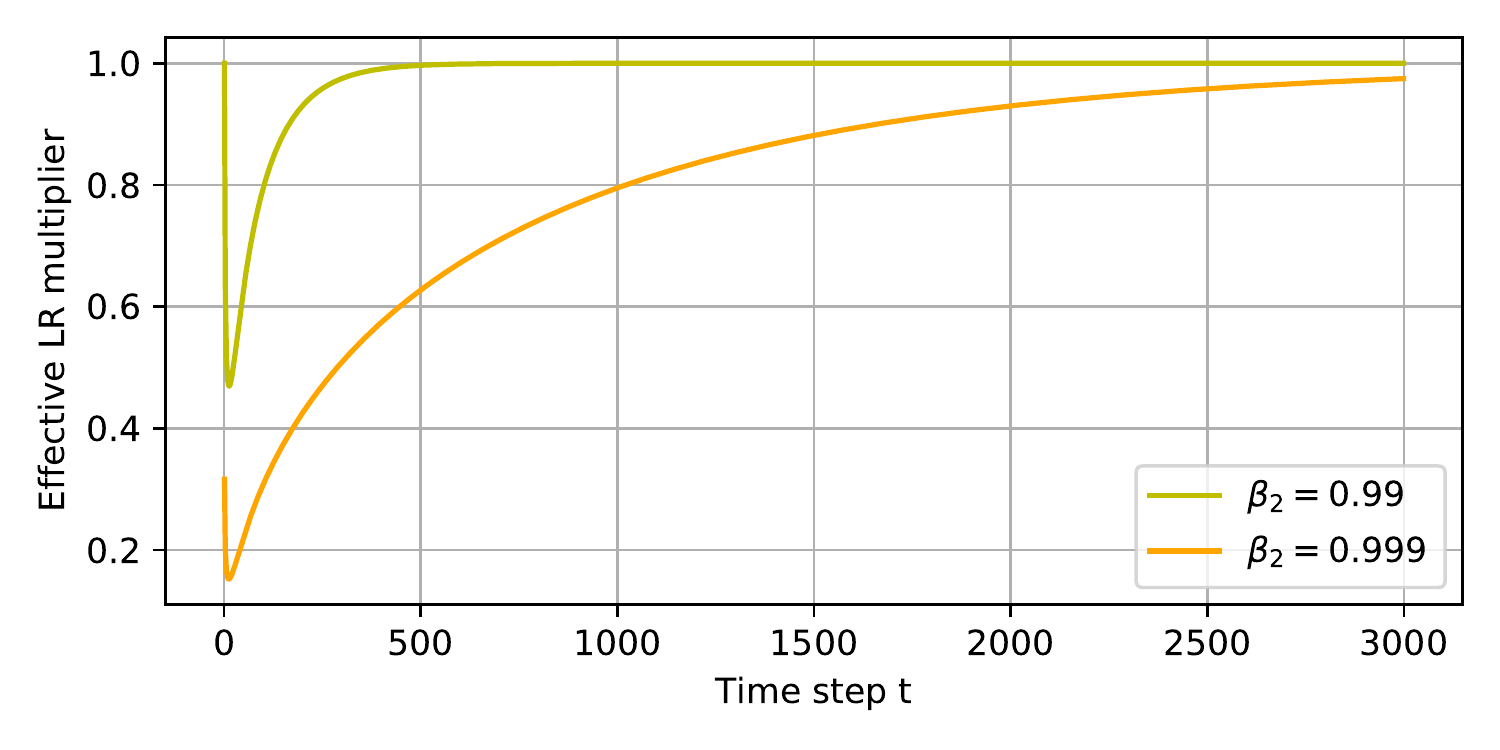}
    \caption{Adam's bias correction factor with $\beta_1 = 0.9$. For common values of $\beta_2$, this takes the shape of a redundant learning rate warmup on top of RMSprop or AdaGrad.}
    \label{fig:biasfactor}
\end{figure}

\paragraph{Warm-up and bias correction.} Adam \cite{kingma2014adam} takes the explicit view of trying to estimate the first and second exponentially-attenuated moments of the gradients. To do so, applies a \emph{bias correction} adjustment factor of $\frac{\sqrt{1 - \beta_2^t}}{1 - \beta_1^t}$ to the learning rate. We note that for common choices of $\beta_1,\beta_2$, this sequence is effectively a learning rate warmup on the scale of $1/(1-\beta_2)$ iterations; see Figure~\ref{fig:biasfactor}.

\paragraph{Accumulator conventions.} In popular implementations, RMSprop and Adam's accumulator updates multiply the new squared gradient by $(1-\beta_2)$; Adam also does this for the momentum $(1-\beta_1)$. This is equivalent to a global scaling of $\eta_t$ and $\eps$ for $\beta_1, \beta_2 < 1$, but prevents a unification with AdaGrad for no fundamental reason. Furthermore, in commonly available implementations of AdaGrad, momentum is not implemented, perhaps due to historical reasons.

\paragraph{The $\bm\eps$ parameter.} This numerical stability hyperparameter, as defined and discussed briefly in Section~\ref{subsec:adaptive-gradients}, may be a hidden confounding factor in experiments. Between frameworks and individual optimizers, the denominator in Algorithm~\ref{alg:generic-grad} may be $\sqrt{m_2 + \eps}$ or $\sqrt{m_2} + \eps$, and default values vary wildly (see Appendix B of \citet{wilson2017marginal}).

We suggest trying $\eps=0$, to remove a confounding factor: if the accumulator $m_2$ is 0, then the movement in that coordinate should also be 0. This can be viewed as taking the square root of the Moore-Penrose pseudoinverse of the accumulator as the preconditioner. In Appendix~\ref{sec:adagrad-noeps-theory}, we verify that this has no bearing on the original AdaGrad regret analysis.

\begin{figure}
    \centering
    \includegraphics[width=\PlotWidth]{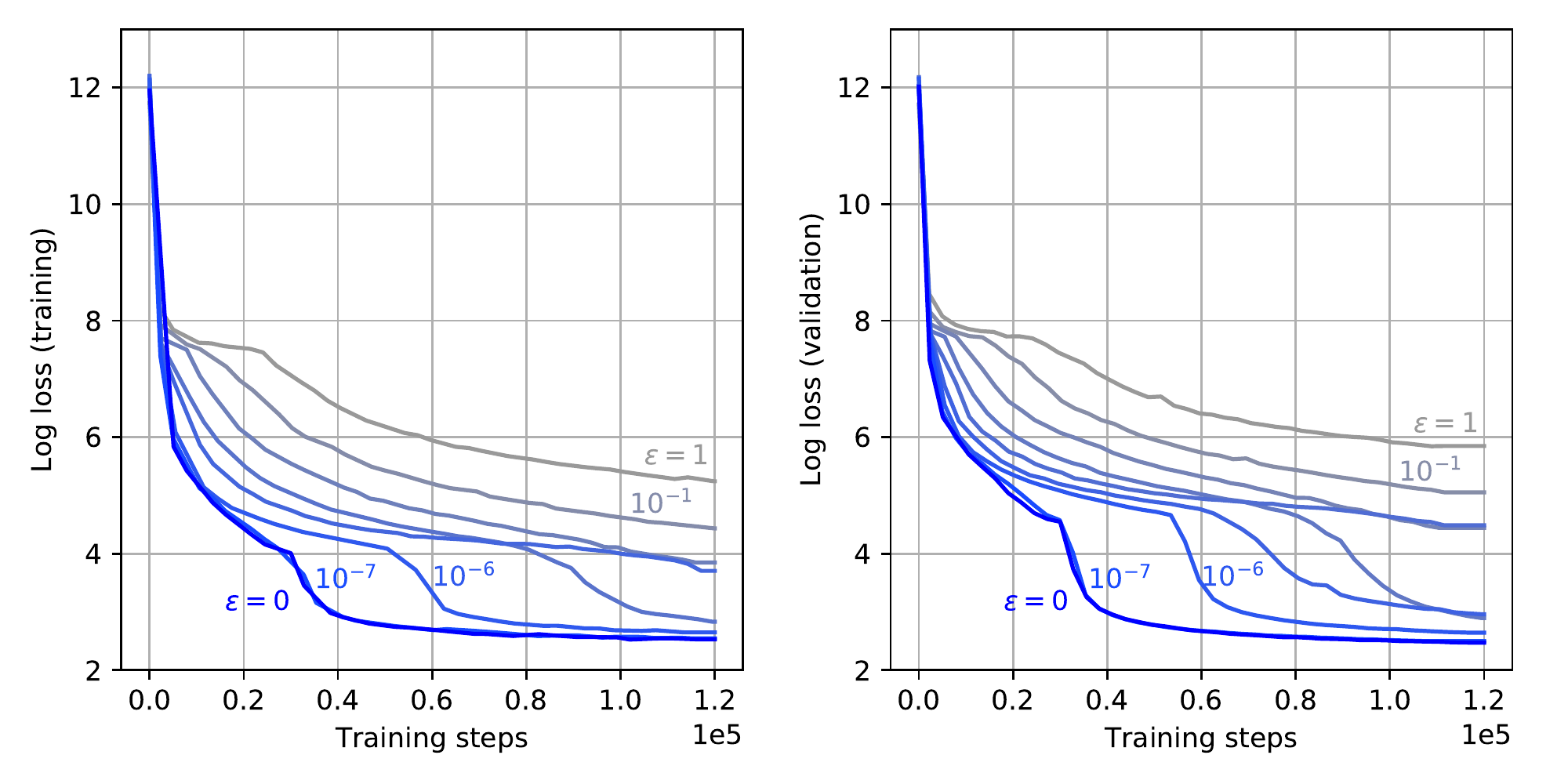}
    \caption{Training curves for AdaGrad in the Transformer machine translation setup of Section~\ref{subsec:wmt}, varying the $\eps$ hyperparameter; this has a dramatic effect on training dynamics. To best demonstrate the interpolation, we did not do any further tuning or grafting.}
    \label{fig:wmt-epsilon}
\end{figure}

Figure~\ref{fig:wmt-epsilon} illustrates how sensitive training can be to the choice of $\eps$, providing a situation where $\eps=0$ is safest.
We do not prescribe this in all situations; dramatically, with a wide-ranging hyperparameter search, \cite{choi2019empirical} settle upon $\eps = 9475$ in one setting.

\subsection{Replicating experiments from \citet{wilson2017marginal}} \label{subsec:replication}

Finally, we report on an attempt to replicate the experimental findings of \citet{wilson2017marginal}, in which they found adaptive methods to generalize poorly in four distinct deep learning setups. Although we were able to replicate some of their findings on these smaller-scale experiments, others appear to be sensitive to hyperparameter tuning, and perhaps subtle changes in the deep learning software and hardware stack that have occurred during the two years since the publication of that paper. We summarize these findings below, deferring some details to Appendix~\ref{subsec:replication-appendix}.

\begin{figure}
    \centering
    \includegraphics[width=\PlotWidth]{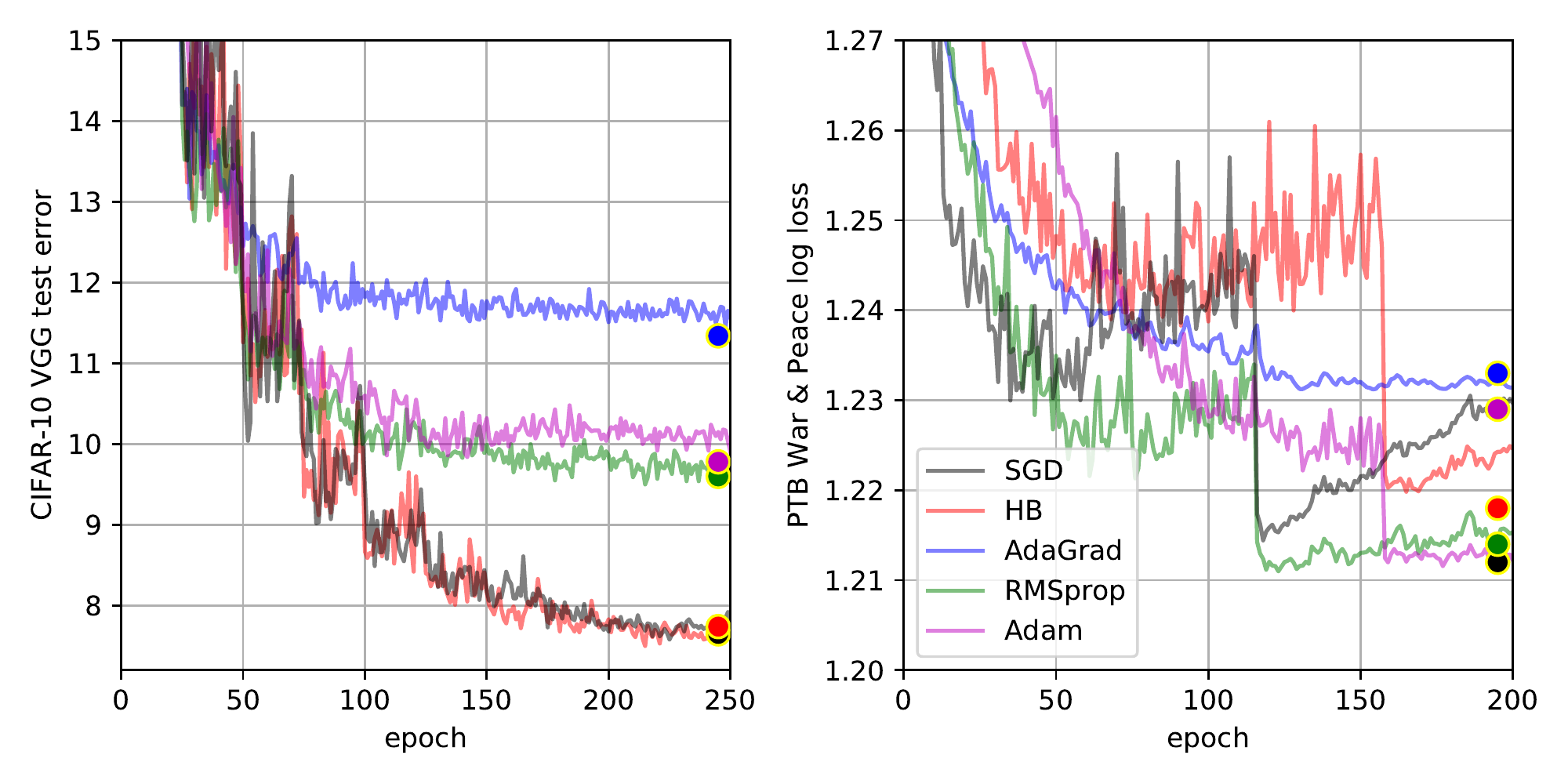}
    \caption{Preview of results of the replication study. Circles show numbers reported by \citet{wilson2017marginal}. In some cases, outdated software/hardware stacks result in mild discrepancies; in others, they may reverse the conclusions.}
    \label{fig:replica-preview}
\end{figure}

For the smaller-scale CIFAR-10 experiment with a VGGNet, we were able to replicate their results more or less perfectly (Figure~\ref{fig:replica-preview}, left). For the character-level language modeling benchmark on War \& Peace, the training setups converged, but suggest the \emph{opposite} conclusion about the generalization of SGD. We had difficulty with the possibly stale codebases for the other two experiments: in the generative parsing experiment, the paper's reported hyperparameters sometimes resulted in non-convergence; even after retuning, we could not match the reported validation losses. We could not get the discriminative parsing experiment, which was written in DyNet and no longer maintained, to finish training before encountering a memory error.

Finally, as an addendum to the CIFAR-10 replication, in Appendix~\ref{subsec:replication-appendix}, we apply a linear learning rate correction (as in Section~\ref{subsec:global-grafting}) to AdaGrad in their setup, resulting in convergence that competes with Adam and RMSprop.

The purpose of this replication attempt, in regards to the thesis of this paper, is to illustrate the unreliability of typical small-scale optimizer comparisons, not only due to the implicit step size schedule; \citet{choi2019empirical} explore this point far more broadly.
\section{Conclusion}
We hope to initiate the practice of learning-rate grafting whenever a new optimizer is introduced. If one observes dramatic differences in the shapes of training curves, we suggest normalizing the implicit step size schedule. This can go both ways:
\begin{itemize}
    \item Grafting a well-performing algorithm $\AlgA$ with $\AlgD :=$ SGD acts as a \emph{sanity check}, determining if $\AlgA$'s superiority comes from finding better step sizes.
    \item Meanwhile, letting $\AlgA$ supply the direction allows one to isolate its preconditioning dynamics, and might aid in the \emph{discovery} of new update rules and schedules.
\end{itemize}
More broadly, we believe that more experiments in optimization for deep learning will benefit from adopting this \emph{bootstrapping} methodology, which depends on tuning one baseline, rather than all candidates. We hope that this will aid in developing more robust beliefs about both adaptive methods and learning rate schedules.

\section*{Acknowledgements}
We are grateful to Sanjeev Arora, Yi Zhang, Zhiyuan Li, Wei Hu, Yoram Singer, Kunal Talwar, Roger Grosse, Karthik Narasimhan, Mark Braverman, Surbhi Goel, and Sham Kakade for helpful discussions and illuminating perspectives.

\bibliography{main}

\newcommand{\etalchar}[1]{$^{#1}$}
\begin{thebibliography}{MKM{\etalchar{+}}94}

\bibitem[ABC{\etalchar{+}}16]{abadi2016tensorflow}
Mart{\'\i}n Abadi, Paul Barham, Jianmin Chen, Zhifeng Chen, Andy Davis, Jeffrey
  Dean, Matthieu Devin, Sanjay Ghemawat, Geoffrey Irving, Michael Isard, et~al.
\newblock Tensorflow: A system for large-scale machine learning.
\newblock In {\em 12th {USENIX} Symposium on Operating Systems Design and
  Implementation}, pages 265--283, 2016.

\bibitem[ABC{\etalchar{+}}19]{pmlr-v97-agarwal19b}
Naman Agarwal, Brian Bullins, Xinyi Chen, Elad Hazan, Karan Singh, Cyril Zhang,
  and Yi~Zhang.
\newblock Efficient full-matrix adaptive regularization.
\newblock In {\em Proceedings of the 36th International Conference on Machine
  Learning}, volume~97 of {\em Proceedings of Machine Learning Research}, pages
  102--110, Long Beach, California, USA, 09--15 Jun 2019. PMLR.

\bibitem[ABH17]{agarwal2017second}
Naman Agarwal, Brian Bullins, and Elad Hazan.
\newblock Second-order stochastic optimization for machine learning in linear
  time.
\newblock {\em The Journal of Machine Learning Research}, 18(1):4148--4187,
  2017.

\bibitem[ADH{\etalchar{+}}19]{arora2019fine}
Sanjeev Arora, Simon~S Du, Wei Hu, Zhiyuan Li, and Ruosong Wang.
\newblock Fine-grained analysis of optimization and generalization for
  overparameterized two-layer neural networks.
\newblock {\em arXiv preprint arXiv:1901.08584}, 2019.

\bibitem[AGK{\etalchar{+}}19]{shampoopp}
Rohan Anil, Vineet Gupta, Tomer Koren, Kevin Regan, and Yoram Singer.
\newblock Full matrix preconditioning made practical.
\newblock {\em Beyond First Order Gradients in Machine Learning Workshop,
  NeurIPS}, 2019.

\bibitem[AGKS19]{anil2019memory}
Rohan Anil, Vineet Gupta, Tomer Koren, and Yoram Singer.
\newblock Memory-efficient adaptive optimization for large-scale learning.
\newblock {\em arXiv preprint arXiv:1901.11150}, 2019.

\bibitem[ALL18]{arora2018theoretical}
Sanjeev Arora, Zhiyuan Li, and Kaifeng Lyu.
\newblock Theoretical analysis of auto rate-tuning by batch normalization.
\newblock {\em arXiv preprint arXiv:1812.03981}, 2018.

\bibitem[AZL19]{allen2019can}
Zeyuan Allen-Zhu and Yuanzhi Li.
\newblock Can sgd learn recurrent neural networks with provable generalization?
\newblock {\em arXiv preprint arXiv:1902.01028}, 2019.

\bibitem[BCN18]{bottou2018optimization}
L{\'e}on Bottou, Frank~E Curtis, and Jorge Nocedal.
\newblock Optimization methods for large-scale machine learning.
\newblock {\em Siam Review}, 60(2):223--311, 2018.

\bibitem[BDS18]{brock2018large}
Andrew Brock, Jeff Donahue, and Karen Simonyan.
\newblock Large scale gan training for high fidelity natural image synthesis.
\newblock {\em arXiv preprint arXiv:1809.11096}, 2018.

\bibitem[BHMM19]{belkin2019reconciling}
Mikhail Belkin, Daniel Hsu, Siyuan Ma, and Soumik Mandal.
\newblock Reconciling modern machine-learning practice and the classical
  bias--variance trade-off.
\newblock {\em Proceedings of the National Academy of Sciences},
  116(32):15849--15854, 2019.

\bibitem[BV04]{boyd2004convex}
Stephen Boyd and Lieven Vandenberghe.
\newblock {\em Convex Optimization}.
\newblock Cambridge University Press, 2004.

\bibitem[BWAA18]{bernstein2018signsgd}
Jeremy Bernstein, Yu-Xiang Wang, Kamyar Azizzadenesheli, and Anima Anandkumar.
\newblock signsgd: Compressed optimisation for non-convex problems.
\newblock {\em arXiv preprint arXiv:1802.04434}, 2018.

\bibitem[BWO18]{blier2018learning}
L{\'e}onard Blier, Pierre Wolinski, and Yann Ollivier.
\newblock Learning with random learning rates.
\newblock {\em arXiv preprint arXiv:1810.01322}, 2018.

\bibitem[BZVL17]{bello2017neural}
Irwan Bello, Barret Zoph, Vijay Vasudevan, and Quoc~V Le.
\newblock Neural optimizer search with reinforcement learning.
\newblock In {\em Proceedings of the 34th International Conference on Machine
  Learning-Volume 70}, pages 459--468, 2017.

\bibitem[CAH{\etalchar{+}}19]{chen2019extreme}
Xinyi Chen, Naman Agarwal, Elad Hazan, Cyril Zhang, and Yi~Zhang.
\newblock Extreme tensoring for low-memory preconditioning.
\newblock {\em arXiv preprint arXiv:1902.04620}, 2019.

\bibitem[CC16]{charniak2016parsing}
Do~Kook Choe and Eugene Charniak.
\newblock Parsing as language modeling.
\newblock In {\em Proceedings of the 2016 Conference on Empirical Methods in
  Natural Language Processing}, pages 2331--2336, 2016.

\bibitem[CG18]{chen2018closing}
Jinghui Chen and Quanquan Gu.
\newblock Closing the generalization gap of adaptive gradient methods in
  training deep neural networks.
\newblock {\em arXiv preprint arXiv:1806.06763}, 2018.

\bibitem[CH16]{cross2016span}
James Cross and Liang Huang.
\newblock Span-based constituency parsing with a structure-label system and
  provably optimal dynamic oracles.
\newblock 2016.

\bibitem[CSN{\etalchar{+}}19]{choi2019empirical}
Dami Choi, Christopher~J Shallue, Zachary Nado, Jaehoon Lee, Chris~J Maddison,
  and George~E Dahl.
\newblock On empirical comparisons of optimizers for deep learning.
\newblock {\em arXiv preprint arXiv:1910.05446}, 2019.

\bibitem[DCLT18]{devlin18}
Jacob Devlin, Ming{-}Wei Chang, Kenton Lee, and Kristina Toutanova.
\newblock {BERT:} pre-training of deep bidirectional transformers for language
  understanding.
\newblock {\em CoRR}, abs/1810.04805, 2018.

\bibitem[DDS{\etalchar{+}}09]{imagenet}
J.~Deng, W.~Dong, R.~Socher, L.-J. Li, K.~Li, and L.~Fei-Fei.
\newblock {ImageNet: A Large-Scale Hierarchical Image Database}.
\newblock In {\em CVPR09}, 2009.

\bibitem[DHS11]{duchi2011adaptive}
John Duchi, Elad Hazan, and Yoram Singer.
\newblock Adaptive subgradient methods for online learning and stochastic
  optimization.
\newblock {\em Journal of Machine Learning Research}, 12(Jul):2121--2159, 2011.

\bibitem[Doz16]{dozat2016incorporating}
Timothy Dozat.
\newblock Incorporating nesterov momentum into adam.
\newblock 2016.

\bibitem[GCH{\etalchar{+}}19]{ginsburg2019stochastic}
Boris Ginsburg, Patrice Castonguay, Oleksii Hrinchuk, Oleksii Kuchaiev, Vitaly
  Lavrukhin, Ryan Leary, Jason Li, Huyen Nguyen, and Jonathan~M Cohen.
\newblock Stochastic gradient methods with layer-wise adaptive moments for
  training of deep networks.
\newblock {\em arXiv preprint arXiv:1905.11286}, 2019.

\bibitem[GDG{\etalchar{+}}17]{goyal2017accurate}
Priya Goyal, Piotr Doll{\'a}r, Ross Girshick, Pieter Noordhuis, Lukasz
  Wesolowski, Aapo Kyrola, Andrew Tulloch, Yangqing Jia, and Kaiming He.
\newblock Accurate, large minibatch sgd: Training imagenet in 1 hour.
\newblock {\em arXiv preprint arXiv:1706.02677}, 2017.

\bibitem[GKKN19]{ge2019step}
Rong Ge, Sham~M Kakade, Rahul Kidambi, and Praneeth Netrapalli.
\newblock The step decay schedule: A near optimal, geometrically decaying
  learning rate procedure for least squares.
\newblock In {\em Advances in Neural Information Processing Systems 32}, pages
  14951--14962. 2019.

\bibitem[GKS18]{gupta2018shampoo}
Vineet Gupta, Tomer Koren, and Yoram Singer.
\newblock Shampoo: Preconditioned stochastic tensor optimization.
\newblock In {\em International Conference on Machine Learning}, 2018.

\bibitem[GKXS18]{gotmare2018closer}
Akhilesh Gotmare, Nitish~Shirish Keskar, Caiming Xiong, and Richard Socher.
\newblock A closer look at deep learning heuristics: Learning rate restarts,
  warmup and distillation.
\newblock {\em arXiv preprint arXiv:1810.13243}, 2018.

\bibitem[GL13]{ghadimi2013stochastic}
Saeed Ghadimi and Guanghui Lan.
\newblock Stochastic first-and zeroth-order methods for nonconvex stochastic
  programming.
\newblock {\em SIAM Journal on Optimization}, 23(4):2341--2368, 2013.

\bibitem[GLSS18a]{gunasekar2018characterizing}
Suriya Gunasekar, Jason Lee, Daniel Soudry, and Nathan Srebro.
\newblock Characterizing implicit bias in terms of optimization geometry.
\newblock {\em arXiv preprint arXiv:1802.08246}, 2018.

\bibitem[GLSS18b]{gunasekar2018implicit}
Suriya Gunasekar, Jason~D Lee, Daniel Soudry, and Nati Srebro.
\newblock Implicit bias of gradient descent on linear convolutional networks.
\newblock In {\em Advances in Neural Information Processing Systems}, pages
  9461--9471, 2018.

\bibitem[Haz16]{hazan2016introduction}
Elad Hazan.
\newblock Introduction to online convex optimization.
\newblock {\em Foundations and Trends{\textregistered} in Optimization},
  2(3-4):157--325, 2016.

\bibitem[HIB{\etalchar{+}}18]{henderson2018deep}
Peter Henderson, Riashat Islam, Philip Bachman, Joelle Pineau, Doina Precup,
  and David Meger.
\newblock Deep reinforcement learning that matters.
\newblock In {\em Thirty-Second AAAI Conference on Artificial Intelligence},
  2018.

\bibitem[HK11]{hazan2011beyond}
Elad Hazan and Satyen Kale.
\newblock Beyond the regret minimization barrier: an optimal algorithm for
  stochastic strongly-convex optimization.
\newblock In {\em Proceedings of the 24th Annual Conference on Learning
  Theory}, pages 421--436, 2011.

\bibitem[HRS15]{hardt2015train}
Moritz Hardt, Benjamin Recht, and Yoram Singer.
\newblock Train faster, generalize better: Stability of stochastic gradient
  descent.
\newblock {\em arXiv preprint arXiv:1509.01240}, 2015.

\bibitem[HZAL18]{haarnoja2018soft}
Tuomas Haarnoja, Aurick Zhou, Pieter Abbeel, and Sergey Levine.
\newblock Soft actor-critic: Off-policy maximum entropy deep reinforcement
  learning with a stochastic actor.
\newblock {\em arXiv preprint arXiv:1801.01290}, 2018.

\bibitem[HZRS16]{he2016identity}
Kaiming He, Xiangyu Zhang, Shaoqing Ren, and Jian Sun.
\newblock Identity mappings in deep residual networks.
\newblock In {\em European Conference on Computer Vision}, pages 630--645.
  Springer, 2016.

\bibitem[IS15]{ioffe2015batch}
Sergey Ioffe and Christian Szegedy.
\newblock Batch normalization: Accelerating deep network training by reducing
  internal covariate shift.
\newblock {\em arXiv preprint arXiv:1502.03167}, 2015.

\bibitem[JYP{\etalchar{+}}17]{jouppi2017datacenter}
Norman~P Jouppi, Cliff Young, Nishant Patil, David Patterson, Gaurav Agrawal,
  Raminder Bajwa, Sarah Bates, Suresh Bhatia, Nan Boden, Al~Borchers, et~al.
\newblock In-datacenter performance analysis of a tensor processing unit.
\newblock In {\em Computer Architecture (ISCA), 2017 ACM/IEEE 44th Annual
  International Symposium on}, pages 1--12. IEEE, 2017.

\bibitem[KALL17]{karras2017progressive}
Tero Karras, Timo Aila, Samuli Laine, and Jaakko Lehtinen.
\newblock Progressive growing of gans for improved quality, stability, and
  variation.
\newblock {\em arXiv preprint arXiv:1710.10196}, 2017.

\bibitem[KB14]{kingma2014adam}
Diederik~P Kingma and Jimmy Ba.
\newblock Adam: A method for stochastic optimization.
\newblock {\em arXiv preprint arXiv:1412.6980}, 2014.

\bibitem[KD18]{kingma2018glow}
Durk~P Kingma and Prafulla Dhariwal.
\newblock Glow: Generative flow with invertible 1x1 convolutions.
\newblock In {\em Advances in Neural Information Processing Systems}, pages
  10215--10224, 2018.

\bibitem[Kri14]{krizhevsky2014one}
Alex Krizhevsky.
\newblock One weird trick for parallelizing convolutional neural networks.
\newblock {\em arXiv preprint arXiv:1404.5997}, 2014.

\bibitem[KS17]{keskar2017improving}
Nitish~Shirish Keskar and Richard Socher.
\newblock Improving generalization performance by switching from adam to sgd.
\newblock {\em arXiv preprint arXiv:1712.07628}, 2017.

\bibitem[LA19]{li2019exponential}
Zhiyuan Li and Sanjeev Arora.
\newblock An exponential learning rate schedule for deep learning.
\newblock {\em arXiv preprint arXiv:1910.07454}, 2019.

\bibitem[LH16]{loshchilov2016sgdr}
Ilya Loshchilov and Frank Hutter.
\newblock Sgdr: Stochastic gradient descent with warm restarts.
\newblock {\em arXiv preprint arXiv:1608.03983}, 2016.

\bibitem[LOG{\etalchar{+}}19]{liu2019roberta}
Yinhan Liu, Myle Ott, Naman Goyal, Jingfei Du, Mandar Joshi, Danqi Chen, Omer
  Levy, Mike Lewis, Luke Zettlemoyer, and Veselin Stoyanov.
\newblock Roberta: A robustly optimized bert pretraining approach.
\newblock {\em arXiv preprint arXiv:1907.11692}, 2019.

\bibitem[MBJ18]{martens2018kronecker}
James Martens, Jimmy Ba, and Matt Johnson.
\newblock Kronecker-factored curvature approximations for recurrent neural
  networks.
\newblock 2018.

\bibitem[MG15]{martens2015optimizing}
James Martens and Roger Grosse.
\newblock Optimizing neural networks with kronecker-factored approximate
  curvature.
\newblock In {\em International {C}onference on {M}achine {L}earning}, pages
  2408--2417, 2015.

\bibitem[MKM{\etalchar{+}}94]{marcus1994penn}
Mitchell Marcus, Grace Kim, Mary~Ann Marcinkiewicz, Robert MacIntyre, Ann Bies,
  Mark Ferguson, Karen Katz, and Britta Schasberger.
\newblock The penn treebank: annotating predicate argument structure.
\newblock In {\em Proceedings of the workshop on Human Language Technology},
  pages 114--119. Association for Computational Linguistics, 1994.

\bibitem[MM19]{mei2019generalization}
Song Mei and Andrea Montanari.
\newblock The generalization error of random features regression: Precise
  asymptotics and double descent curve.
\newblock {\em arXiv preprint arXiv:1908.05355}, 2019.

\bibitem[MS10]{mcmahan2010adaptive}
H~Brendan McMahan and Matthew Streeter.
\newblock Adaptive bound optimization for online convex optimization.
\newblock {\em arXiv preprint arXiv:1002.4908}, 2010.

\bibitem[NDG{\etalchar{+}}17]{dynet}
Graham Neubig, Chris Dyer, Yoav Goldberg, Austin Matthews, Waleed Ammar,
  Antonios Anastasopoulos, Miguel Ballesteros, David Chiang, Daniel Clothiaux,
  Trevor Cohn, Kevin Duh, Manaal Faruqui, Cynthia Gan, Dan Garrette, Yangfeng
  Ji, Lingpeng Kong, Adhiguna Kuncoro, Gaurav Kumar, Chaitanya Malaviya, Paul
  Michel, Yusuke Oda, Matthew Richardson, Naomi Saphra, Swabha Swayamdipta, and
  Pengcheng Yin.
\newblock Dynet: The dynamic neural network toolkit.
\newblock {\em arXiv preprint arXiv:1701.03980}, 2017.

\bibitem[Nes83]{nesterov1983method}
Yurii~Evgen'evich Nesterov.
\newblock A method of solving a convex programming problem with convergence
  rate o(k\^{}2).
\newblock In {\em Doklady Akademii Nauk}, volume 269, pages 543--547. Russian
  Academy of Sciences, 1983.

\bibitem[Nes13]{nesterov2013introductory}
Yurii Nesterov.
\newblock {\em Introductory lectures on convex optimization: A basic course},
  volume~87.
\newblock Springer Science \& Business Media, 2013.

\bibitem[PB18]{popel2018training}
Martin Popel and Ond{\v{r}}ej Bojar.
\newblock Training tips for the transformer model.
\newblock {\em The Prague Bulletin of Mathematical Linguistics}, 110(1):43--70,
  2018.

\bibitem[PMB13]{pascanu2013difficulty}
Razvan Pascanu, Tomas Mikolov, and Yoshua Bengio.
\newblock On the difficulty of training recurrent neural networks.
\newblock In {\em International conference on machine learning}, pages
  1310--1318, 2013.

\bibitem[Pol64]{polyak1964some}
Boris~T Polyak.
\newblock Some methods of speeding up the convergence of iteration methods.
\newblock {\em USSR Computational Mathematics and Mathematical Physics},
  4(5):1--17, 1964.

\bibitem[RKK18]{reddi2018convergence}
Sashank~J Reddi, Satyen Kale, and Sanjiv Kumar.
\newblock On the convergence of {A}dam and beyond.
\newblock In {\em International Conference on Learning Representations}, 2018.

\bibitem[SBH19]{schneider2019deepobs}
Frank Schneider, Lukas Balles, and Philipp Hennig.
\newblock Deepobs: A deep learning optimizer benchmark suite.
\newblock {\em arXiv preprint arXiv:1903.05499}, 2019.

\bibitem[SNW{\etalchar{+}}19]{shen2019lingvo}
Jonathan Shen, Patrick Nguyen, Yonghui Wu, Zhifeng Chen, et~al.
\newblock Lingvo: a modular and scalable framework for sequence-to-sequence
  modeling, 2019.

\bibitem[SRK{\etalchar{+}}19]{staib2019escaping}
Matthew Staib, Sashank~J Reddi, Satyen Kale, Sanjiv Kumar, and Suvrit Sra.
\newblock Escaping saddle points with adaptive gradient methods.
\newblock {\em arXiv preprint arXiv:1901.09149}, 2019.

\bibitem[SS18]{shazeer2018adafactor}
Noam Shazeer and Mitchell Stern.
\newblock Adafactor: Adaptive learning rates with sublinear memory cost.
\newblock {\em arXiv preprint arXiv:1804.04235}, 2018.

\bibitem[ST19]{smith2019super}
Leslie~N Smith and Nicholay Topin.
\newblock Super-convergence: Very fast training of neural networks using large
  learning rates.
\newblock In {\em Artificial Intelligence and Machine Learning for Multi-Domain
  Operations Applications}, volume 11006, page 1100612. International Society
  for Optics and Photonics, 2019.

\bibitem[SZ14]{simonyan2014very}
Karen Simonyan and Andrew Zisserman.
\newblock Very deep convolutional networks for large-scale image recognition.
\newblock {\em arXiv preprint arXiv:1409.1556}, 2014.

\bibitem[TH12]{tieleman2012lecture}
Tijmen Tieleman and Geoffrey Hinton.
\newblock Lecture 6.5-rmsprop: Divide the gradient by a running average of its
  recent magnitude.
\newblock {\em COURSERA: Neural networks for machine learning}, 4(2):26--31,
  2012.

\bibitem[VSP{\etalchar{+}}17]{vaswani2017attention}
Ashish Vaswani, Noam Shazeer, Niki Parmar, Jakob Uszkoreit, Llion Jones,
  Aidan~N Gomez, {\L}ukasz Kaiser, and Illia Polosukhin.
\newblock Attention is all you need.
\newblock In {\em Advances in Neural Information Processing Systems}, pages
  5998--6008, 2017.

\bibitem[WRS{\etalchar{+}}17]{wilson2017marginal}
Ashia~C Wilson, Rebecca Roelofs, Mitchell Stern, Nati Srebro, and Benjamin
  Recht.
\newblock The marginal value of adaptive gradient methods in machine learning.
\newblock In {\em Advances in Neural Information Processing Systems}, pages
  4151--4161, 2017.

\bibitem[WXY{\etalchar{+}}17]{wen2017terngrad}
Wei Wen, Cong Xu, Feng Yan, Chunpeng Wu, Yandan Wang, Yiran Chen, and Hai Li.
\newblock Terngrad: Ternary gradients to reduce communication in distributed
  deep learning.
\newblock In {\em Advances in neural information processing systems}, pages
  1509--1519, 2017.

\bibitem[YDY{\etalchar{+}}19]{yang2019xlnet}
Zhilin Yang, Zihang Dai, Yiming Yang, Jaime Carbonell, Ruslan Salakhutdinov,
  and Quoc~V Le.
\newblock Xlnet: Generalized autoregressive pretraining for language
  understanding.
\newblock {\em arXiv preprint arXiv:1906.08237}, 2019.

\bibitem[YGG17]{you2017large}
Yang You, Igor Gitman, and Boris Ginsburg.
\newblock Large batch training of convolutional networks.
\newblock {\em arXiv preprint arXiv:1708.03888}, 2017.

\bibitem[YLH{\etalchar{+}}19]{you2019reducing}
Yang You, Jing Li, Jonathan Hseu, Xiaodan Song, James Demmel, and Cho-Jui
  Hsieh.
\newblock Reducing bert pre-training time from 3 days to 76 minutes.
\newblock {\em arXiv preprint arXiv:1904.00962}, 2019.

\bibitem[Zei12]{zeiler2012adadelta}
Matthew~D Zeiler.
\newblock Adadelta: an adaptive learning rate method.
\newblock {\em arXiv preprint arXiv:1212.5701}, 2012.

\bibitem[Zin03]{zinkevich2003online}
Martin Zinkevich.
\newblock Online convex programming and generalized infinitesimal gradient
  ascent.
\newblock In {\em Proceedings of the 20th International Conference on Machine
  Learning (ICML-03)}, pages 928--936, 2003.

\bibitem[ZRS{\etalchar{+}}18]{zaheer2018adaptive}
Manzil Zaheer, Sashank Reddi, Devendra Sachan, Satyen Kale, and Sanjiv Kumar.
\newblock Adaptive methods for nonconvex optimization.
\newblock In {\em Advances in neural information processing systems}, pages
  9793--9803, 2018.

\end{thebibliography}
\bibliographystyle{alpha}

\newpage
\appendix
\onecolumn

\section{Summary of Empirical Results}
\label{appendix:experiment-guide}

For convenience, we provide a summary of all of the assorted experiments, which are interleaved with discussion and definitions in the main paper. Pointers to the relevant sections, figures, and tables accompany each main point.

\paragraph{Layer-wise grafting: training curves depend strongly on $\AlgM$.} In both the ResNet and Transformer settings, training curves and final performance metrics depend significantly more heavily on $\AlgM$, the sub-algorithm in AdaGraft supplying the magnitude. No additional tuning was done on the grafted optimizers. This is our primary result. SGD remains the best-performing optimizer on ResNet (by a smaller margin), but the $\AlgM =$ AdaGrad grafting instances perform \emph{better} than plain AdaGrad. References: Sections~\ref{subsec:imagenet} and \ref{subsec:wmt}, Figures~\ref{fig:imagenet-graft} and Figure~\ref{fig:wmt-graft}, and Tables~\ref{table:imagenet-results} and \ref{table:wmt-results}. Hyperparameter details are in Appendix~\ref{appendix:experiment-details}.

\paragraph{Successfully training a Transformer with (grafted) SGD.} A minor side note is that it has been an empirical challenge to train a state-of-the-art Transformer NLP model using SGD; the empirical papers we have cited use Adam and AdaGrad, and \cite{you2019reducing} corroborate this in passing (motivating their LAMB optimizer as an improvement over LARS). We show that SGD can train such a model if it is allowed to take bootstrapped layer-wise learning rates from an algorithm that trains successfully. Completing the story and training with a aptly-corrected pure SGD is an interesting line of future work.

\paragraph{Global grafting: training curves depend on $\AlgM$ in vision but not NLP.} In the ResNet setting, the same result as layer-wise grafting is seen with global grafting, but the global learning rate (one scalar) needs to be decreased slightly (factor of 0.5). This weaker, secondary result is shown for only $\AlgM =$ SGD, $\AlgD =$ AdaGrad. Global grafting did \emph{not} work in the Transformer setting, even upon tuning the global learning rate multiplier. References: Section~\ref{subsec:global-grafting} and \ref{subsec:wmt-transfer}, Figures~\ref{fig:imagenet-transfer} and \ref{fig:wmt-transfer}.

\begin{figure}[H]
    \centering
    \includegraphics[width=\AppendixPlotWidth]{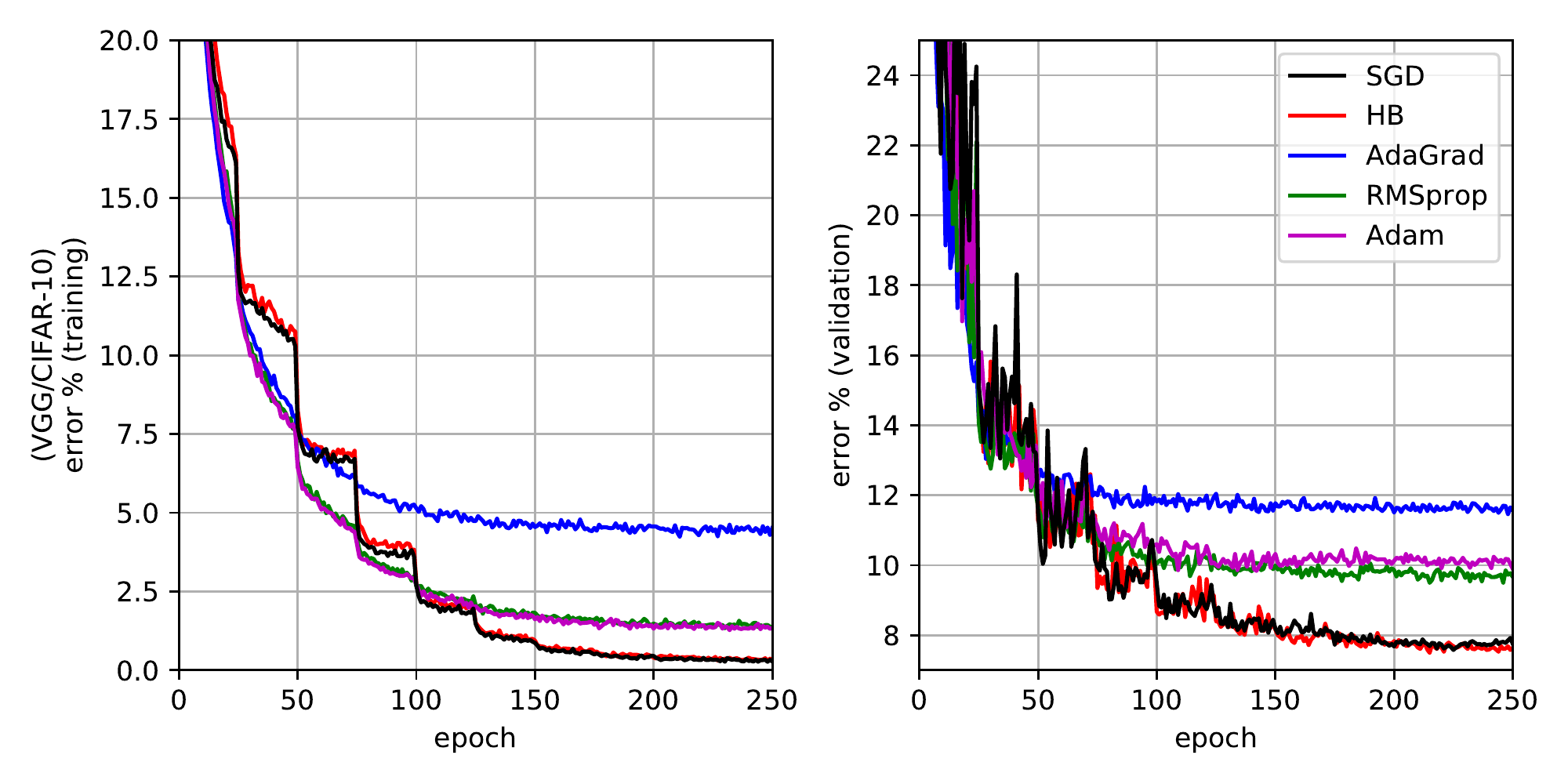}
    \caption{Replication of CIFAR-10 image classification experiment from \cite{wilson2017marginal}. Results were consistent with those reported in the paper.}
    \label{fig:replica-cifar}
\end{figure}

\paragraph{Linear schedule for AdaGrad on image models.} From the global grafting experiments, we can find a (purely empirical) correction for AdaGrad's step sizes: a linearly increasing schedule $(c_0 + c_1 t)$, to be multiplied on top of the existing schedule $\eta_t$. Also shown to work on a VGGNet on CIFAR-10, as a side note. References: Section~\ref{subsec:global-grafting}, Figures~\ref{fig:imagenet-transfer} and \ref{fig:replica-cifar}.

\paragraph{Isolating the preconditioning behavior of recently proposed optimizers.} Letting $\AlgD$ be Yogi, Adafactor, or AMSGrad, but fixing $\AlgM =$ AdaGrad, we get nearly indistinguishable training curves in the Transformer setting. If these optimizers are performing well on their own (ungrafted), it is because they are finding good layer-wise step sizes. References: Section~\ref{subsec:implications}, Figure~\ref{fig:wmt-exotic}.

\paragraph{Varying the $\eps$ parameter.} In the Transformer setting, tuning $\eps$ can have a dramatic effect on convergence. In this case, below $10^{-7}$, the curves are the same. Thus, we suggest trying ``$\eps = 0$'' to simplify the hyperparameter search space. References: Section~\ref{subsec:software}, Figure~\ref{fig:wmt-epsilon}.

\paragraph{Exploratory visualization of layerwise step corrections.} A visualization that sheds light on why adaptive optimizer choice is brittler in the Transformer setting than the ResNet setting. For the ResNet, the scalar corrections implied by \emph{layer-wise} grafting roughly match that of global grafting; they are all well-approximated by the linear correction heuristic. For the Transformer, these sequences can differ by up to 5 orders of magnitude. In particular, AdaGrad takes hundreds of times larger steps on the softmax and embedding layers, compared to SGD. Also, the layer-wise correction schedules have high-dimensional structure at the beginning of training. References: Section~\ref{subsec:nlp}, Figure~\ref{fig:all-corr}.

\begin{figure}[H]
    \centering
    \includegraphics[width=\AppendixPlotWidth]{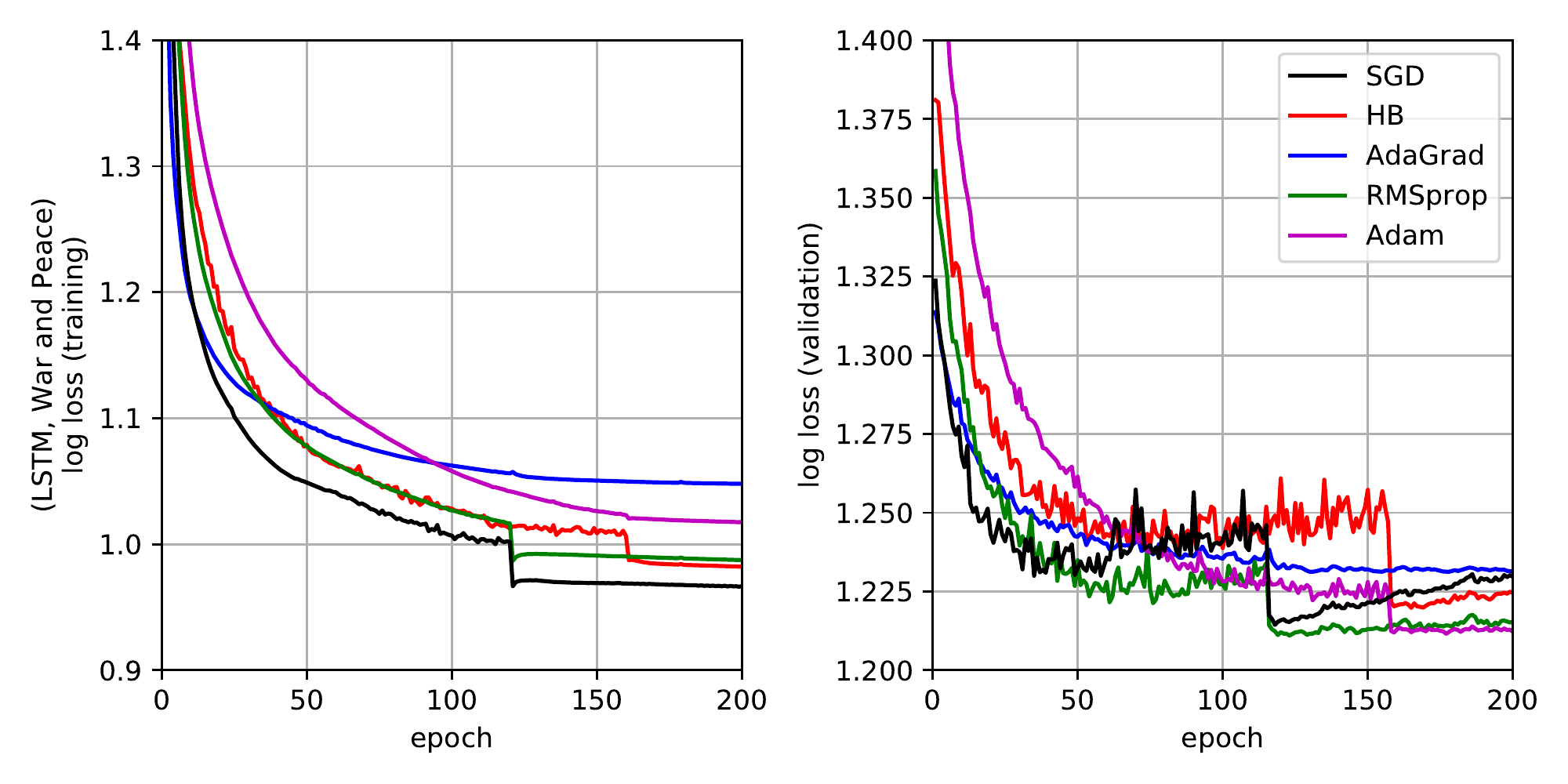}
    \caption{Failed replication of War \& Peace character-level language modeling experiment from \cite{wilson2017marginal}. Although every model converged, results were not consistent with the paper's findings.}
    \label{fig:replica-lm}
\end{figure}

\paragraph{Replications of \citet{wilson2017marginal} experiments.} We could reproduce the CIFAR-10 experiments, but not the other three. Note that our main ImageNet experiments challenge the suggestion from the CIFAR-10 experiments that adaptive methods are insufficient for image models, and that the linear correction to AdaGrad transfers to this setting. References: Appendix~\ref{subsec:replication-appendix}, Figures~\ref{fig:replica-cifar}, \ref{fig:replica-lm}, \ref{fig:replica-parser}, Tables~\ref{table:replica-cifar}, \ref{table:replica-lm}, \ref{table:replica-parser}.
\section{Supplemental Figures and Tables} \label{appendix:supp-figures}

\subsection{Global grafting and schedule correction for machine translation} \label{subsec:wmt-transfer}
\begin{figure}[H]
    \centering
    \includegraphics[width=\AppendixPlotWidth]{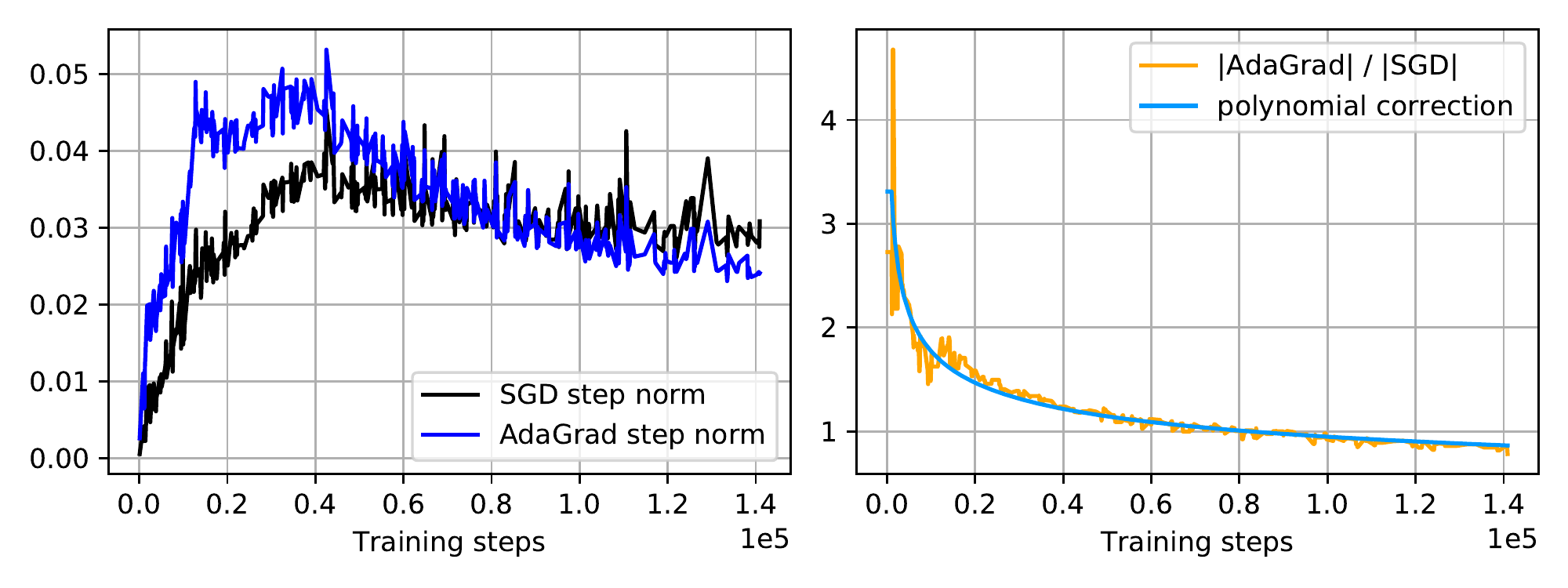}
    \caption{Step norm visualization, as in Figure~\ref{fig:imagenet-stepnorms}, for the Transformer/WMT setup, layer-wise grafting with $\AlgM$=AdaGrad, $\AlgD$=SGD. The global norm ratio suggests an offline learning rate schedule correction proportional to $t^\alpha$ for $\alpha \approx -0.27$, but applying this does not cause SGD to train well. }
    \label{fig:wmt-stepnorms}
\end{figure}

We show the negative result we encountered when applying the learning rate schedule bootstrapping experiment of Section~\ref{subsec:global-grafting} to the machine translation model. We begin by repeating the procedure of discovering a learning rate schedule correction; as seen in Figure~\ref{fig:wmt-stepnorms}, this results in a near-perfect fit for an inverse polynomial schedule correction, with power approximately 0.27.

However, this learning rate correction, despite tuning, did not result in improved training; see Figure~\ref{fig:wmt-transfer} in the main paper. Even after tuning the global scalar learning rate multiplier, we could not get SGD with the $t^{-0.27}$ learning rate schedule correction to match the training curve of AdaGrad. Only \emph{layer-wise grafted} AdaGrad could match (in fact, outperform) AdaGrad itself. In Section~\ref{sec:discussion}, we develop more fine-grained empirical evidence for why one might expect learning rate schedule tuning to be more challenging in natural language processing than in vision.

\subsection{Replication of \citet{wilson2017marginal} experiments} \label{subsec:replication-appendix}


\paragraph{CIFAR-10: positive replication.} On the classic benchmark task of CIFAR-10 classification with a VGG network \citep{simonyan2014very}, we were able to replicate the \citep{wilson2017marginal} results perfectly, using the same codebase\footnote{\url{https://github.com/szagoruyko/cifar.torch}}. We repeated the hyperparameter search reported in the paper, found the same optimal base learning rates for each optimizer, and found the same stratification in performance between non-adaptive methods, Adam \& RMSprop, and AdaGrad. \citet{choi2019empirical} have scrutinized the same experiment at greater depth, testing the effects of tuning.

\begin{table}[H]
\centering
\begin{tabular}{|c|c|c|}
\hline
\textbf{Optimizer} & \textbf{Test error (original)} & \textbf{Test error (replication)} \\ \hline
\textbf{SGD} & $7.65 \pm 0.14$ (1) & $7.76 \pm 0.06$ (1) \\ \hline
\textbf{HB} & $7.74 \pm 0.25$ (2) & $7.64 \pm 0.03$ (2) \\ \hline
\textbf{AdaGrad} & $11.34 \pm 0.46$ (5) & $11.57 \pm 0.09$ (5) \\ \hline
\textbf{RMSprop} & $9.60 \pm 0.19$ (3) & $9.67 \pm 0.12$ (3) \\ \hline
\textbf{Adam} & $9.78 \pm 0.25$ (4) & $9.90 \pm 0.11$ (4) \\ \hline
\textbf{AdaGrad (linear correction)} & --- & $8.92 \pm 0.10$ \\ \hline
\end{tabular}
\caption{Data for replication of CIFAR-10 experiments. Standard deviations are computed over 5 trials. The linear schedule correction discovered in Section~\ref{subsec:global-grafting}, with $(c_0,c_1) = (0,10^{-8})$, improves AdaGrad's convergence significantly.}
\label{table:replica-cifar}
\end{table}


\paragraph{Char-RNN: negative replication.} Curiously, our replication of the language modeling experiment using the same popular repository\footnote{\url{https://github.com/jcjohnson/torch-rnn}} was successful in reproducing the optimal hyperparameter settings, but resulted in the \emph{opposite} conclusion. Here, SGD found the objective with the smallest training loss, but Adam exhibited the best generalization performance. We believe that software version discrepancies (our setup: CUDA 10.1, cuDNN 7.5.1) may account for these differences.

\begin{table}[H]
\centering
\begin{tabular}{|c|c|c|}
\hline
\textbf{Optimizer} & \textbf{Val loss (original)} & \textbf{Val loss (replication)} \\ \hline
\textbf{SGD} & $1.212 \pm 0.001$ (1) & $1.230 \pm 0.004$ (4) \\ \hline
\textbf{HB} & $1.218 \pm 0.002$ (3) & $1.224 \pm 0.002$ (3) \\ \hline
\textbf{AdaGrad} & $1.233 \pm 0.004$ (5) & $1.231 \pm 0.002$ (5) \\ \hline
\textbf{RMSprop} & $1.214 \pm 0.005$ (2) & $1.215 \pm 0.002$ (2) \\ \hline
\textbf{Adam} & $1.229 \pm 0.004$ (4) & $1.213 \pm 0.003$ (1) \\ \hline
\end{tabular}
\caption{Data for replication of War \& Peace experiments. Standard deviations are computed over 5 trials.}
\label{table:replica-lm}
\end{table}

\begin{figure}[H]
    \centering
    \includegraphics[width=\AppendixPlotWidth]{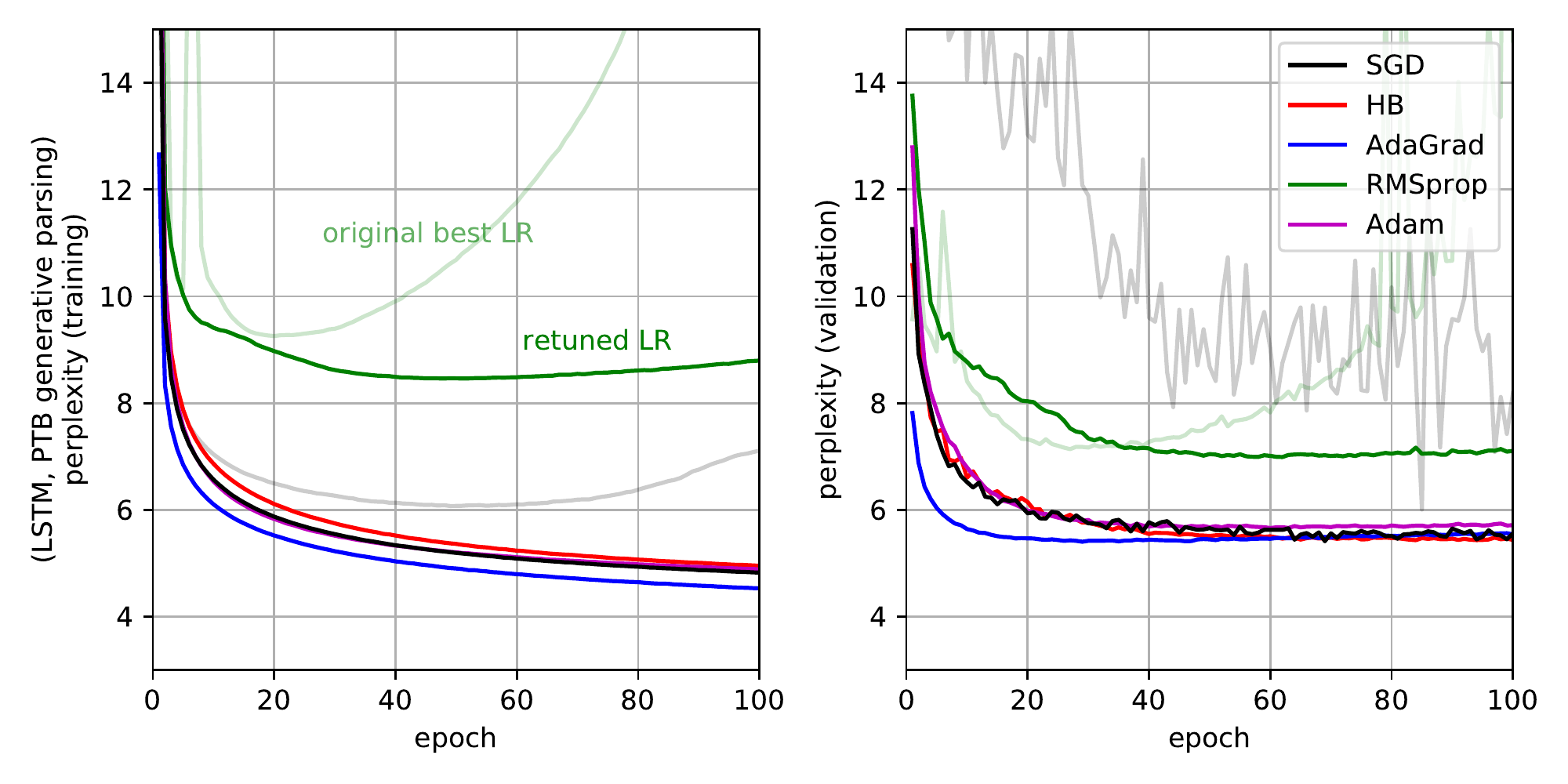}
    \caption{Failed replication of generative parsing experiments. Convergence seems to be very unstable; even after retuning hyperparameters for the wayward optimizers, results were not consistent with the paper's findings.}
    \label{fig:replica-parser}
\end{figure}

\paragraph{PTB generative parsing: positive replication.} Next, for the experiments on the Penn Treebank \citep{marcus1994penn} constituency parsing code\footnote{\url{https://github.com/cdg720/emnlp2016}} accompanying \citep{charniak2016parsing}, using the same architectural and training protocol modifications as specified in \citep{wilson2017marginal}, we were able to get the model to converge with each optimizer. However, for two of the settings (SGD and RMSprop), the best reported learning rates exhibited non-convergence (the fainter curves in Figure~\ref{fig:replica-parser}). Similarly as the above experiment, the ranking of optimizers' training and generalization performance differs from that seen in the original report.

\begin{table}[H]
\centering
\begin{tabular}{|c|c|c|}
\hline
\textbf{Optimizer} & \textbf{Val ppl. (original)} & \textbf{Val ppl. (replication)} \\ \hline
\textbf{SGD} & $5.09 \pm 0.04$ (1) & $5.53 \pm 0.05$ (2) \\ \hline
\textbf{HB} & $5.13 \pm 0.01$ (2) & $5.46 \pm 0.01$ (1) \\ \hline
\textbf{AdaGrad} & $5.24 \pm 0.02$ (3) & $5.56 \pm 0.01$ (3) \\ \hline
\textbf{RMSprop} & $5.28 \pm 0.00$ (4) & $7.11 \pm 0.02$ (5) \\ \hline
\textbf{Adam} & $5.35 \pm 0.01$ (5) & $5.72 \pm 0.01$ (4) \\ \hline
\end{tabular}
\caption{Data for replication of Penn Treebank generative parsing experiments. Standard deviations are computed over 5 trials.}
\label{table:replica-parser}
\end{table}

\paragraph{PTB discriminative parsing: could not run.} Finally, \cite{wilson2017marginal} include a fourth set of experiments, generative parsing of Penn Treebank, using the code\footnote{\url{https://github.com/jhcross/span-parser}} accompanying \citep{cross2016span}. Unfortunately, this DyNet \citep{dynet} implementation, which was last updated in 2016, encountered a fatal memory leak when training with our DyNet 2.1 setup. Along the same lines as the random-seed-tuning experiments of \cite{henderson2018deep}, this suggests that there are further technical complications to the problems of credible optimizer evaluation addressed by \cite{schneider2019deepobs}, even on well-known supervised learning benchmarks.
\section{Experiment Details} \label{appendix:experiment-details}
\subsection{Efficient implementation of AdaGraft}
\label{subsec:implementation}
We provide some notes on implementation concerns when applying AdaGraft to large-scale settings.

\paragraph{Black-box grafting.} In popular software packages, due to efficiency concerns, an optimizer implementation will often update a variable \emph{in-place}. For sake of completeness, we show how to do this with some auxiliary space in Algorithm~\ref{alg:adagraft-impl}. Of course, this is not necessary when grafting two known optimizers; in that case, the operation can be streamlined significantly by computing the closed-form per-coordinate steps directly.

\begin{algorithm} 
\caption{ AdaGraft, black-box implementation }
\label{alg:adagraft-impl}
\begin{algorithmic}[1]
\STATE \textbf{Input: } Optimizers $\AlgM, \AlgD$; initializer $x_1$; $\eps > 0$.
\STATE Initialize $\AlgM, \AlgD$ at $x_1$.
\FOR{$t = 1, \ldots, T$}
  \STATE Receive stochastic gradient $g_t$.
  \STATE Save $w_t$ to scratch space $s_t$.
  \STATE Run $\AlgM$ in-place: $w_t \leftarrow \AlgM(w_t, g_t)$.
  \STATE Compute step norm: $\eta_\AlgM := \norm{w_t - s_t}$.
  \STATE Reload weights from scratch space: $w_t \leftarrow s_t$.
  \STATE Run $\AlgD$ in-place: $w_t \leftarrow \AlgD(w_t, g_t)$.
  \STATE Compute step norm: $\eta_\AlgD := \norm{w_t - s_t}$.
  \STATE Update: $w_{t+1} \leftarrow s_t + \frac{\eta_\AlgM}{\eta_\AlgD + \eps}(w_t - s_t)$.
\ENDFOR
\end{algorithmic}
\end{algorithm}

\paragraph{Numerical considerations.} Although the global step size scalar supplied to $\AlgD$ can be arbitrary (it never appears in the), when using in-place operations, numerical precision can be lost when the update is much smaller than the $w_t$ itself. We suggest using a multiplier of $1$ to minimize this problem; choosing scales like $10^{-4}$ causes grafting to be numerically unstable. This is especially pertinent with huge models trained using mixed-precision floating point numbers.

\paragraph{Parallelism.} We note that global grafting can be significantly slower in realistic large-scale training pipelines, since it is harder to parallelize. In particular, the grafted step magnitudes depend on a round of synchronization after computing all of the per-layer gradients. Meanwhile, distributed training frameworks provide per-parameter abstractions which naturally support the straightforward implementation of per-layer grafting.

\subsection{ImageNet classification} \label{subsec:imagenet-details}
The ResNet-50 models were trained with batch size 4096, with $\ell_2$ regularization $10^{-4}$ and label smoothing 0.1.

For the base (ungrafted) optimizers, we follows a staircase learning rate schedule where learning rate is ramped up linearly from 0 to 6.4 over the first 5 epochs, followed by $10\times$ drop in learning rate at 30 epochs, 60 epochs and 80 epochs. Momentum is set to $0.9$ throughout. For Adam, we use the default $\beta_2 = 0.999$. We used the $\eps = 0$ rule we introduced for the adaptive optimizers. Global learning rate scalars for SGD, Adam, and AdaGrad are set to $10^{-1}$, $10^{-4}$, and $10^{-3}$, respectively; these were found via coarse grid search. In fact, global learning rate was the only hyperparameter we tuned separately; we derived the rest from the adaptive optimizers' defaults, along with a well-tuned SGD setup. Importantly, note that no experimental conclusion depends on the premise that an optimizer setup's hyperparameters have been tuned to the same degree as another's (which is very hard to fulfill or certify).

\subsection{WMT14 English-French translation} \label{subsec:wmt-details}
The Transformer models were trained with batch size 384, and residual, input, attention, and ReLU dropout 0.1. The architecture hyperparameters are as follows: hidden state dimension 8192, 32K subword encoding tokens, embedding dimension 1024, and 12 attention heads.

For all optimizers, we used a learning rate schedule with a linear warm-up over 40K steps. For SGD, we kept the learning rate constant at the peak; for AdaGrad and Adam, we used a $1/\sqrt{t}$ decay like in \cite{shazeer2018adafactor}. Like in the ImageNet experiments, momentum is set to $0.9$ throughout, Adam's $\beta_2 = 0.999$, $\eps = 0$ rule was used for the adaptive methods. Again via grid search, we found global learning rates of 0.03, 0.015, and 0.03 for SGD, Adam, and AdaGrad, respectively.

\section{Proofs} \label{appendix:proofs}
\subsection{Pathological constructions} \label{subsec:pathological-theory}

In this section we provide two simple examples of stochastic convex problems where it can be seen that when it comes to generalization both AdaGrad and SGD can be significantly better than the other depending on the instance. Our purpose to provide both the examples is to stress our point that the issue of understanding the generalization performance of SGD vs. adaptive methods is more nuanced than what simple examples might suggest and hence such examples should be treated as qualitative indicators more for the purpose of providing intuition. Indeed which algorithm will perform better on a given problem, depends on various properties of the precise instance.  

\paragraph{Example where SGD $>$ AdaGrad.}
We provide a brief intuitive review of the construction provided by \cite{wilson2017marginal}; for a precise description, see Section~3.3 of that paper. Consider a setting of overparameterized linear regression, where the true output (i.e. dependent variable) $y\in\{\pm 1\}$ is the first coordinate of the feature vector (independent variable) $x$. The next two coordinates of $x$ are ``dummy'' coordinates set to 1; then, the coordinates are arranged in blocks which only appear once per sample, taking the value of $y$.

The key idea is that in this setting, the solution space that AdaGrad explores is always in the subspace of the sign vector of $X^{\top}y$. As a result, AdaGrad treats the first three coordinates essentially indistinguishably putting equal mass on each. It can then be seen that for any new example the AdaGrad solution does not extract the true label information from the first three coordinates and hence gets the prediction wrong, leading to high generalization error; the other distinguishing features belong to the new unique block which are set to 0 for the AdaGrad solution, as it has not seen them.

\paragraph{Example where AdaGrad $>$ SGD.}
This example is motivated from the original AdaGrad paper \citep{duchi2011adaptive}, adapted to the overparameterized setting. Consider a distribution $\mathcal{Z}$ supported over $\{0,1\}^d$ with equal $1/d$ mass over vectors with exactly one $1$ and $0$ mass everywhere else. Let the label distribution be always $y=1$. Consider sampling a dataset $S$ of size $c\cdot d$ where $c \leq 1$ (corresponding to the overparameterized setting) and consider the hinge loss
\[f_t(x) = [1 - y_t(z_t^{\top}x_t)]_{+}\]
where $(z_t,y_t)$ denotes the $t$-th (example, label) pair. Note that there is an optimal predictor given by the all-ones vector. 

Running AdaGrad in such a setting, it can be seen that the first time a vector that has not appeared yet is sampled, AdaGrad quickly adapts by setting the coordinate corresponding to the vector to $1$ and thereby making $0$ error on the example. Therefore after one epoch of AdaGrad ($cd$ steps), the training error reduces to $0$ and the average test error becomes roughly $~ (1 - c)$. On the other hand, for SGD (with an optimal $1/\sqrt{t}$ decay scheme) after say $cd/2$ steps, the learning rate reduces to at most $O(1/\sqrt{d})$ and therefore in the next $cd/2$ steps SGD reduces the error at most by a factor of $O(1 - 1\sqrt{d})$, leading to a total test error of at least $\sim\!(1 - c/2)$ after a total of $cd$ steps. This is significantly smaller than the error achieved by AdaGrad at this stage. Further note that to get down to the same test error as that achieved by AdaGrad, it can be seen that SGD requires at least $\Omega(\sqrt{d})$ times more steps than AdaGrad.

\subsection{Regret of AdaGrad with $\eps = 0$} \label{subsec:eps-zero}
\label{sec:adagrad-noeps-theory}

In Section \ref{subsec:software}, we give a method for setting $\eps$ to 0, to ensure that an adaptive optimizer is not interpolating with SGD. We give a brief note on why $\eps$ is unnecessary to establish the regret bound from \cite{duchi2011adaptive}. The general update is performed as follows:
\[
    w_{t+1} = w_t - \eta~H_t^{\dagger}g_t;
    \quad
    G_{t+1} = G_t + g_t g_t\tr,
\]
where the preconditioning matrix $H_t$ is updated in two possible ways:
\[\text{\textbf{diagonal}: } H_{t} \defeq \text{diag}(G_t)^{1/2}; \qquad \text{\textbf{full}: }H_{t} \defeq G_t^{1/2}.\]

The following theorem whose proof is very similar to the original analysis in \cite{duchi2011adaptive} shows that the above modification leads to no change in the regret guarantee of AdaGrad. 
\begin{theorem}
The regret of AdaGrad, with pseudoinverse updates, is bounded as
\begin{align*}
    \text{\emph{(full)}\quad} 
    \text{Regret} &\leq O\left( \max_{t \leq T} \|w_t - w^*\|_2 \; \mathrm{tr}(G_T^{1/2})\right) \\
    \text{\emph{(diag)}\quad} 
    \text{Regret} &\leq O\left( \max_{t \leq T} \|w_t - w^*\|_{\infty} \; \sum_{i=1}^d \sqrt{\sum_{t=1}^T g_{t,i}^2}\right)
\end{align*}
\end{theorem}

We now provide a quick proof sketch for the above highlighting the main parts of the proof that change from the standard version. As in the original proof we consider the case of linear loss functions at every step given by $g_t$. The first step is to note that the following relationship holds directly by the definition of the updates:
\begin{align*}
&g_t^\top (x_t - x^*) \\
&\qquad= \frac{1}{2\eta}(\|x_{t} - x^*\|_{H_t}^2 - \|x_{t+1} - x^*\|_{H_t}^2) \\
&\qquad+ \frac{\eta}{2}(g_t^{\top} H_t^{\dagger}H_tH_t^{\dagger} g_t)^2.
\end{align*}
The analysis now follows in the standard way by summing the above over time and analyzing the first and second summation separately. The first term is the same as the standard analysis and therefore leads to no change. 

Further more for the second term the idea of the original proof is to show that
\[\sum_{t} (g_t^{\top} H_t^{\dagger}H_tH_t^{\dagger} g_t)^2 \leq 2 \cdot \mathrm{tr}(H_T).\]
The above statement follows in the same way as the original proof with care for pseudoinverses. For instance in the diagonal version we can apply Lemma 4 from \cite{duchi2011adaptive} along each coordinate separately applying the lemma from the first time the coordinate sees a non-zero gradient and ignoring everything before as it is 0. 

\end{document}